\documentclass[final]{cvpr}

\usepackage{times}
\usepackage{epsfig}
\usepackage{graphicx}
\usepackage{amsmath}
\usepackage{amssymb}

\usepackage[pagebackref=true,breaklinks=true,colorlinks,bookmarks=false]{hyperref}

\usepackage{amsmath,amsfonts,bm}

\def\eqref#1{equation~\ref{#1}}

\def\1{\bm{1}}

\def\vmu{{\bm{\mu}}}

\def\vb{{\bm{b}}}

\def\vg{{\bm{g}}}

\def\vi{{\bm{i}}}

\def\vk{{\bm{k}}}

\def\vo{{\bm{o}}}

\def\vs{{\bm{s}}}

\def\vx{{\bm{x}}}
\def\vy{{\bm{y}}}
\def\vz{{\bm{z}}}

\def\mA{{\bm{A}}}

\def\mI{{\bm{I}}}

\def\mW{{\bm{W}}}
\def\mX{{\bm{X}}}

\def\mSigma{{\bm{\Sigma}}}

\DeclareMathAlphabet{\mathsfit}{\encodingdefault}{\sfdefault}{m}{sl}
\SetMathAlphabet{\mathsfit}{bold}{\encodingdefault}{\sfdefault}{bx}{n}

\usepackage{cleveref}
\usepackage{csquotes}
\usepackage{amsmath}
\usepackage{amssymb}
\usepackage{hyperref}
\usepackage{bm}
\usepackage[acronym,nomain,shortcuts]{glossaries}
\usepackage{url}
\usepackage{layouts}
\usepackage{enumitem}
\usepackage{diagbox}
\usepackage{multirow}
\usepackage{hhline}
\usepackage{booktabs}
\usepackage{makecell}
\usepackage{siunitx}
\usepackage{comment}
\glsdisablehyper

\let\oldsqrt\sqrt
\def\sqrt{\mathpalette\DHLhksqrt}
\def\DHLhksqrt#1#2{%
  \setbox0=\hbox{$#1\oldsqrt{#2\,}$}\dimen0=\ht0
  \advance\dimen0-0.2\ht0
  \setbox2=\hbox{\vrule height\ht0 depth -\dimen0}%
  {\box0\lower0.4pt\box2}}

\crefname{section}{Sec.}{Secs.}
\crefname{subsection}{Sec.}{Secs.}
\crefname{subsubsection}{Sec.}{Secs.}
\crefname{paragraph}{Par.}{Para.}
\crefname{appendix}{App.}{Apps.}
\crefname{table}{Tab.}{Tabs.}
\crefname{figure}{Fig.}{Figs.}
\crefname{equation}{Eq.}{Eq.}
\Crefname{equation}{Eq.}{Eq.}
\crefname{algorithm}{Alg.}{Algs.}

\newcommand{\mnist}[1]{\raisebox{-2px}{\includegraphics[width=14px,height=14px]{figs/mnist/#1}}}
\newcommand{\fmnist}[1]{\raisebox{-2px}{\includegraphics[width=14px,height=14px]{figs/fashionmnist/#1}}}
\newcommand{\devan}[1]{\raisebox{-2px}{\includegraphics[width=16px,height=16px]{figs/devanagari/#1}}}

\newcommand{\vbeta}{{\bm{\beta}}}
\newcommand{\vgamma}{{\bm{\gamma}}}
\newcommand{\vpi}{{\bm{\pi}}}

\newcommand{\parheading}[1]{\vspace{0.5em}\par\noindent\textbf{#1}\hspace{.5em}}

\mathchardef\mhyphen="2D

\def\cvprPaperID{16} 
\def\confYear{CLVISION 2021}

\begin{document}

\title{Continual Learning with Fully Probabilistic Models}

\author{Benedikt Pf{\"u}lb, Alexander Gepperth and Benedikt Bagus\\
Fulda University of Applied Sciences\\
Fulda, Germany\\
{\tt\small \{benedikt.pfuelb,alexander.gepperth,benedikt.bagus\}@cs.hs-fulda.de}
}

\maketitle
\begin{abstract}
\noindent We present an approach for continual learning (CL) that is based on fully probabilistic (or generative) models of machine learning. 
In contrast to, e.g., GANs being \enquote{generative} in the sense that they can generate samples, fully probabilistic models that aim at modeling the data distribution directly. 
Consequently, they provide functionalities that are highly relevant for continual learning, such as density estimation (outlier detection) and sample generation. 
As a concrete realization of generative continual learning, we propose Gaussian Mixture Replay (GMR).
GMR is a pseudo-rehearsal approach using a Gaussian Mixture Model (GMM) instance for both, generator and classifier functionalities.
Relying on the  MNIST, FashionMNIST and Devanagari benchmarks, we demonstrate unsupervised task boundary detection by GMM density estimation, which we also use to reject untypical generated samples. 
In addition, we show that GMR is capable of class-conditional sampling in the way of a cGAN.
Lastly, we verify that GMR, despite its simple structure, achieves state-of-the-art performance on common class-incremental learning problems at very competitive time and memory complexity. 
\end{abstract}
\section{Introduction}\label{sec:intro}
\parheading{Context} 
This conceptual work is in the context of continual learning (CL). 
In its most general formulation, CL assumes that the distribution of training data changes over the training time of a machine learning model (concept drift).
Often, this is restricted to a succession of sub-tasks having a stable data distribution, with abrupt changes in data distribution occurring at \textit{sub-task boundaries} only. 
This is what we refer to as Sequential Learning Task (SLT), see \cref{sec:SLT}.
\par 
The CL paradigm is completely agnostic w.r.t.\ the type of learning that is involved.
Most current work on CL is about supervised learning, often in the context of classification which usually requires discriminative machine learning methods. 
Since such methods are not well-suited for \textit{outlier detection}, the recognition of sub-task boundaries is problematic. 
The problem is usually circumvented by simply assuming that sub-task boundaries are known.
\par 
In contrast to conventional machine learning, learning in CL occurs over long times.
This implies a number of constraints. 
First, access to data is limited, typically to samples from the current sub-task due to memory reasons.
Of course, a small subset of samples from previous sub-tasks may be retained. 
The \textit{generation} of such samples is more useful.
Second, training times for new sub-tasks should scale sub-linearly (ideally $\mathcal{O}(1)$) with the total number of samples seen by the model. 
Otherwise, CL could not be scaled to learning tasks with an infinite number of sub-tasks.
\parheading{Motivation} 
The presented work is motivated by the fact that many functionalities evoked in the previous paragraphs are typical for generative, unsupervised learning methods. 
Mixture models, for example, are commonly used for outlier detection and sample generation.
They have a benign forgetting behavior when faced with changes in data distribution. 
In this article, we aim at integrating mixture models into a hybrid approach for supervised CL, which we term Gaussian Mixture Replay (GMR).
Moreover, we want to show the various benefits for CL on standard benchmarks. 
\subsection{Related Work on CL}\label{sec:relwork}
\begin{figure}[htb!]
	\centering
	\includegraphics[width=\linewidth]{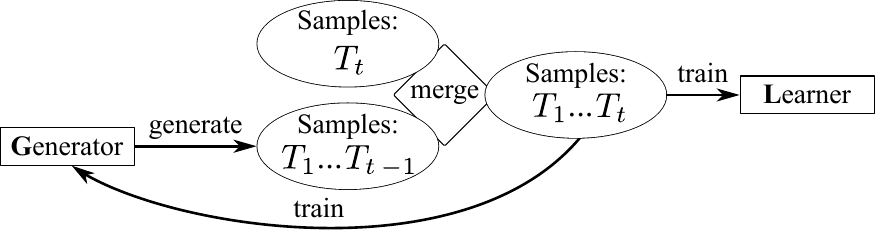}
	\caption{
		The replay approach to continual learning: a \textbf{L}earner, e.g., a DNN, is trained on several sub-tasks sequentially.
		To avoid forgetting, a \textbf{G}enerator is trained to generate samples from past sub-tasks. 
		For training \textbf{L}, \textbf{G} \textit{generates} samples from past sub-tasks, which are merged with current sub-task samples.
	}
	\label{fig:replay}
\end{figure}

\noindent The field of CL is expanding rapidly, see \cite{Parisi2018,Hayes2018,Soltoggio2017,Lange2019} for reviews.
Systematic comparisons between different approaches to avoid CF are performed in, e.g., \cite{Kemker2017,Pfuelb2019}.
As discussed in \cite{Pfuelb2019}, many recently proposed methods demand specific experimental setups, which deviate significantly from application scenarios.
For example, some methods require access to samples from \textit{all} sub-tasks for tuning hyper-parameters, whereas others need access to all samples from past sub-tasks.
Many proposed methods have a time and/or memory complexity that scales at least linear with the number of sub-tasks and, thus, may fail if this number is large.
Among the proposed remedies to CF, three major directions may be distinguished according to \cite{Lange2019}: parameter isolation, regularization and rehearsal.
\parheading{Parameter Isolation} 
Isolation methods aim at determining (or creating) a group of DNN parameters that are mainly \enquote{responsible} for a certain sub-task. 
CL is then avoided by \textit{protecting} these parameters when training on successive sub-tasks. 
Representative works are \cite{Fernando2017,Mallya2017,Mallya2018,Serra2018,Rusu2016,Aljundi2016}.
\parheading{Regularization} 
Regularization methods usually propose the modification of the loss function, including additional terms that protect knowledge acquired in previous sub-tasks.
Current approaches are very diverse: SSL \cite{Aljundi2018} focuses on enhancing sparsity of neural activities, whereas approaches such as LwF \cite{Li2016} rely on knowledge distillation mechanisms.
A method that has attracted significant attention is Elastic Weight Consolidation (EWC) \cite{Kirkpatrick2016}.
EWC inhibits changes to weights that are important to previous sub-tasks, while measuring this importance based on the Fisher information matrix (FIM).
An online variant of EWC is published~\cite{Schwarz2018}.
Synaptic intelligence \cite{Zenke2017} is pursuing a similar goal.
Incremental Moment Matching \cite{Lee2017} makes use of the FIM to merge the parameters obtained for different sub-tasks. 
The Matrix of Squares (MasQ) method \cite{Gepperth2019SIM} is similar to EWC, but relies on the calculus of derivatives to assess the importance of parameters for a sub-task.
It is more simple w.r.t.\ its concepts and more memory-efficient.
\parheading{Rehearsal}
Two main methods can be distinguished: rehearsal and pseudo-rehearsal.
\textit{Rehearsal} methods store a subset of samples from past sub-tasks preventing CF, either by putting constraints on current sub-task training or by adding retained samples to the current sub-task training set.
Typical representatives of rehearsal methods are iCaRL \cite{Rebuffi2016}, (A-)GEM \cite{LopezPaz2017, Chaudhry2018}, GBSS \cite{Aljundi2019} and TEM \cite{Chaudhry2019}. 
\textit{Pseudo-rehearsal} or \textit{generative replay} methods, in contrast, do not store samples but generate them using a dedicated \textit{generator} that is trained along with the learner, see \cref{fig:replay}.
Typical models used as generators are Generative Adversarial Networks (GANs), Variational AutoEncoders (VAEs) and their variants, see \cite{Shin2017} and \cite{Kamra2017}. 
The GMR model we are proposing belongs to the latter type as well.
\parheading{Training and Evaluation Paradigms for CL} 
In the context of CL, a wide range of training and evaluation paradigms are proposed, see \cite{Buzzega2020,Joseph2020,We2019,Farquhar2018,Kemker2017,Lesort2018,Mundt2020}.
\subsection{Gaussian Mixture Replay}\label{sec:gmr_intro}
\begin{figure}[htb!]
	\centering
	\includegraphics[width=\linewidth]{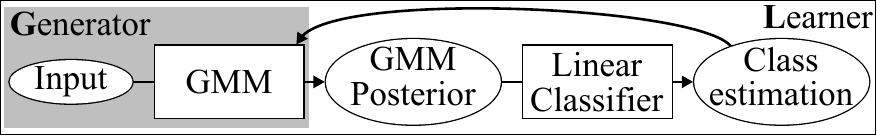}
	\caption{
		Principal structure of the GMR model, composed of a GMM modeling the distribution of training samples (left).
		A linear classifier is  operating on the posterior probabilities (also termed \textit{responsibilities}) produced by the GMM. 
		The coupled GMM/classifier implements the \textbf{L}earner, whereas the GMM implements the \textbf{G}enerator from \cref{fig:replay}.
		The GMM sampling process is informed by feedback from the classifier. 
	}
	\label{fig:gmr}
\end{figure}

\noindent Gaussian Mixture Replay (GMR) is a CL approach based on pseudo-rehearsal with a Gaussian Mixture Model (GMM) serving as generator. 
Mixture models describe the probability density of data $\mX$ as a weighted superposition of parametric distributions $p(\cdot;\vbeta_j)$: 
\begin{equation*}
	p(\mX) = \prod_i p(\vx_i) = \prod_i \sum_{j=1}^K \pi_j p(\vx_i;\vbeta_j).
\end{equation*}
For GMR, we use Gaussian parametric distributions defined by centroids $\vmu_j$ and covariance matrices $\mSigma_j$: \mbox{$p(\vx;\vbeta_j)$\,$\equiv$\,$\mathcal{N}(\vx ; \vmu_j, \mSigma_j)$\,$\equiv$\,$\mathcal N_j(\vx)$}. 
For simplicity, we describe GMR including a single GMM \enquote{layer} only, but a generalization to deep convolutional GMMs is straightforward, see \cite{gepperth2021a}. 
Data vectors entering the trained GMM are transformed into the GMM's a posteriori distibution (or \textit {responsibility}) $\vgamma$ as $\gamma_i(\vx) = \frac{\exp(\mathcal{N}_i(\vx))}{\sum_z \exp(\mathcal{N}_z(\vx))}$. 
Responsibilities are bound in the interval $[0,1]$ and normalized to have unit sum: $\sum_z \gamma_z(\vx)$\,$=$\,$1$. 
This makes them well suited as inputs for a linear classifier which transforms responsibilities into class membership probabilities. 
The data flow through a GMR instance is shown in \cref{fig:gmr}.
\par
A major advantage of GMR is that the generator and learner are not separate entities. 
The GMM performs generative tasks (sampling and outlier detection), and, at the same time, provides the learner (i.e., the linear classifier) with a high-level data representation. 
\subsection{Differences to Related Work}
\noindent Gaussian Mixture Replay (GMR) aims to improve the following aspects of recent work on continual learning:
\parheading{Outlier Detection} 
Discriminative machine learning models such as DNNs or CNNs, which are at the heart of most current CL approaches, allow \textit{supervised} outlier detection only. 
Here, outliers are simply samples with high losses, and concept drift is assumed to occur if the loss changes significantly. 
However, loss computation requires targets for supervised learning, which are not always available.
Problematic in such an approach, however, that is impossible to determine whether concept drift is occurring in the data, or if it is just the targets drifting.
Furthermore, outlier detection for individual samples cannot be trusted: high-loss inliers cannot be distinguished from outliers, unless classification accuracy is almost perfect.
\parheading{Sample Generation} 
In pseudeo-rehearsal methods such as \cite{Shin2017}, GANs (cGANs, WGANs) are employed as generators.
While these can generate impressive samples, it is not clear whether these samples represent the full probability distribution that they are supposed to sample from. 
In fact, there is the problem of \textit{mode collapse}, where GANs focus on a small part of the data distribution only. 
Mode collapse is difficult to detect automatically, since GANs do not possess a (differentiable) loss function expressing the models' current ability to sample.
\parheading{Resource Efficiency} 
Pseudo-rehearsal approaches contain generator and learner components. 
For GANs, the generator is composed of not only a generator, but also a discriminator. 
These components are usually implemented as DNNs or CNNs requiring a considerable amount of resources, particularly memory. 
\parheading{Scalability} 
Since the generators are implemented as CNNs or DNNs, they are very sensitive to class balance. 
Therefore, the generator must produce the precise number of samples for each new sub-task so that classes from previous and current sub-tasks are balanced.
As a consequence, the number of generated samples grows linearly with the number of sub-tasks, which may be unsustainable for problems with many sub-tasks.
\subsection{Novel Contributions}
\noindent GMR offers several novel contributions to the field of CL: 
\begin{itemize}[leftmargin=*,nosep]
	\setlength\itemsep{0em}
	\item Unsupervised outlier detection: consistency ensured by relying on a fully probabilistic GMM
	\item Resource-efficiency: pseudo-rehearsal integrating learner and generator in a single structure
    \item Robustness: model collapse excluded by theoretical guarantees for GMM training
	\item Competitiveness: state-of-the-art CL performance on standard problems
\end{itemize}
In order to validate our approach, we perform a comparison to the Elastic Weight Consolidation (EWC) model.
That is assumed to be a \enquote{standard model} for CL in many recent publications.
A simple generative-replay approach as presented in \cite{Shin2017} is used as baseline.
Furthermore, we provide a public TensorFlow 2 implementation\footnote{\scriptsize\url{https://gitlab.cs.hs-fulda.de/ML-Projects/clwfpm}}.
\section{Data}
\parheading{Image Benchmarks}\label{sec:datasets}
In order to measure the impact of forgetting during continual learning, three public image classification benchmarks are used to construct sequential learning tasks, see \cref{tab:datasets}.
All datasets consist of grayscale images with dimensions of $28$\,$\times$\,$28$\,$\times$\,$1$ or $32$\,$\times$\,$32$\,$\times$\,$1$, whose entries are normalized to the $[0,1]$ interval.
We merge the provided train and test sets for each benchmark, and split the merged data in a proportion of $90\%$ to $10\%$ into training and test data. 
All datasets exhibit an almost equal distribution of samples within classes.
\par
MNIST contains images of handwritten digits ($0$-$9$) with a resolution of $28$\,$\times$\,$28$ pixels.
It is probably the most commonly used benchmark for classification problems.
FashionMNIST contains pictures of different types of clothes.
This dataset is supposed to be harder to classify compared to MNIST (same resolution) and, thus, leads to lower accuracies.
Similar to MNIST, the Devanagari dataset contains written Devanagari letters. 
It is available in a resolution of $32$\,$\times$\,$32$ pixels per image.
As this dataset include more classes than needed, we randomly select $10$ classes.
\parheading{Sequential Learning Tasks}\label{sec:SLT}
Sequential Learning Tasks (SLTs) simulate a continuous learning scenario by dividing datasets given in \cref{sec:datasets}.
The resulting sub-datasets are enumerated and contain only samples of non-overlapping classes.
For example, a $D_{5\text{-}5}$ task consists of two sub-datasets consisting of $5$ classes each.
Every sub-task is identified by its order, e.g., $T_1$, $T_2$, $\ldots$, $T_x$.
Baseline experiments ($D_{10}$) contain all available classes to investigate the effect of incremental task-by-task training.
\par
With SLTs basic experiments can be carried out to determine the effect of forgetting under the conditions above.
To measure the impact of the number of classes contained in a task, different combinations and subdivisions are evaluated.
\Cref{tab:SLTs} displays all evaluated SLTs and their definition of sub-tasks.
\begin{table}[h!]
	\centering
	\caption{Definition of Sequential Learning Tasks (SLTs) and the class divisions of their sub-tasks.
	} 
	\label{tab:SLTs}
	\setlength\tabcolsep{3pt}
	\begin{tabular}{|l|l|}
		\hline
		\textbf{SLT}                                 & \textbf{Sub-Tasks}                                                    \\ \hline
		$D_{10}$                                     & $T_1($0, 1, 2, 3, 4, 5, 6, 7, 8, 9$)$                                 \\ \hline
		$D_{9\text{-}1a}$                            & $T_1($0, 1, 2, 3, 4, 5, 6, 7, 8$)$~~~$T_2($9$)$                       \\ \hline
		$D_{9\text{-}1b}$                            & $T_1($0, 1, 2, 4, 5, 6, 7, 8, 9$)$~~~$T_2($3$)$                       \\ \hline
		$D_{5\text{-}5a}$                            & $T_1($0, 1, 2, 3, 4$)$~~~$T_2($5, 6, 7, 8, 9$)$                       \\ \hline
		$D_{5\text{-}5b}$                            & $T_1($0, 1, 2, 6, 7$)$~~~$T_2($3, 4, 5, 8, 9$)$                       \\ \hline
		$D_{2\text{-}2\text{-}2\text{-}2\text{-}2a}$ & $T_1($0, 1$)$~$T_2($2, 3$)$~$T_3($4, 5$)$~$T_4($6, 7$)$~$T_5($8, 9$)$ \\ \hline
		$D_{2\text{-}2\text{-}2\text{-}2\text{-}2b}$ & $T_1($1, 7$)$~$T_2($0, 2$)$~$T_3($6, 8$)$~$T_4($4, 5$)$~$T_5($3, 9$)$ \\ \hline
	\end{tabular}
\end{table}
\begin{table*}
	\newcolumntype{a}{>{\centering\arraybackslash}m{16px}}
	\centering
	\caption{Detailed information to the used datasets (including examples of each class).}
	\label{tab:datasets}
	\setlength\tabcolsep{2pt}
	\begin{tabular}{|l|c|c|c|c|aaaaaaaaaa|}
		\hline
		\multicolumn{1}{|c|}{\textbf{Dataset}} &   \textbf{Ref.}    & \textbf{Resolution}  &    \textbf{Number of}     &  \textbf{Number of}   &                                  \multicolumn{10}{c|}{Random \textbf{Examples} (from classes)}                                  \\
		                                       &                    &                      & \textbf{Training Samples} & \textbf{Test Samples} & 0          & 1          & 2          & 3          & 4          & 5          & 6          & 7          & 8          & 9          \\ \hline
		MNIST                                  &  \cite{Lecun1998}  & $28$\,$\times$\,$28$ &          50\,000          &        10\,000        & \mnist{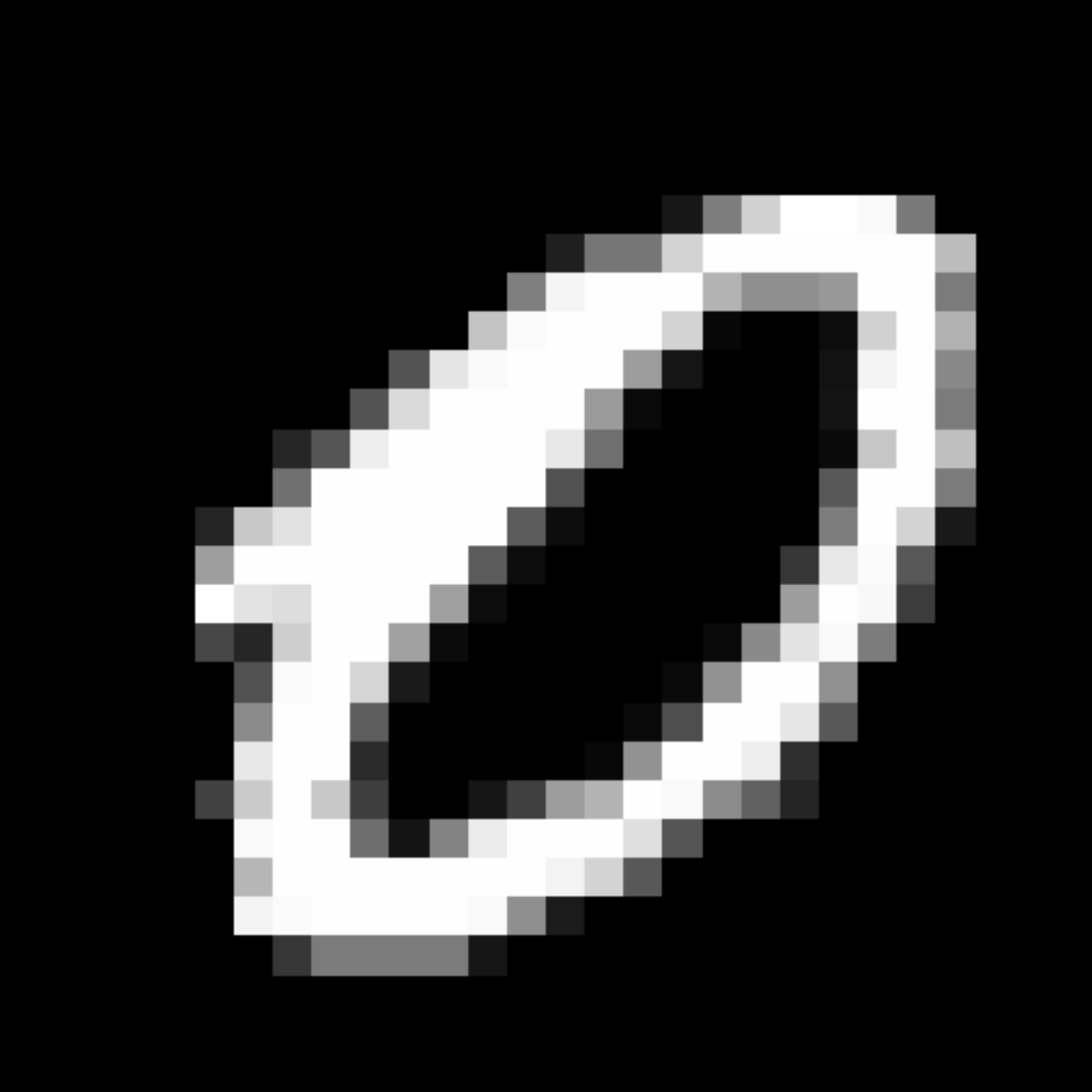}  & \mnist{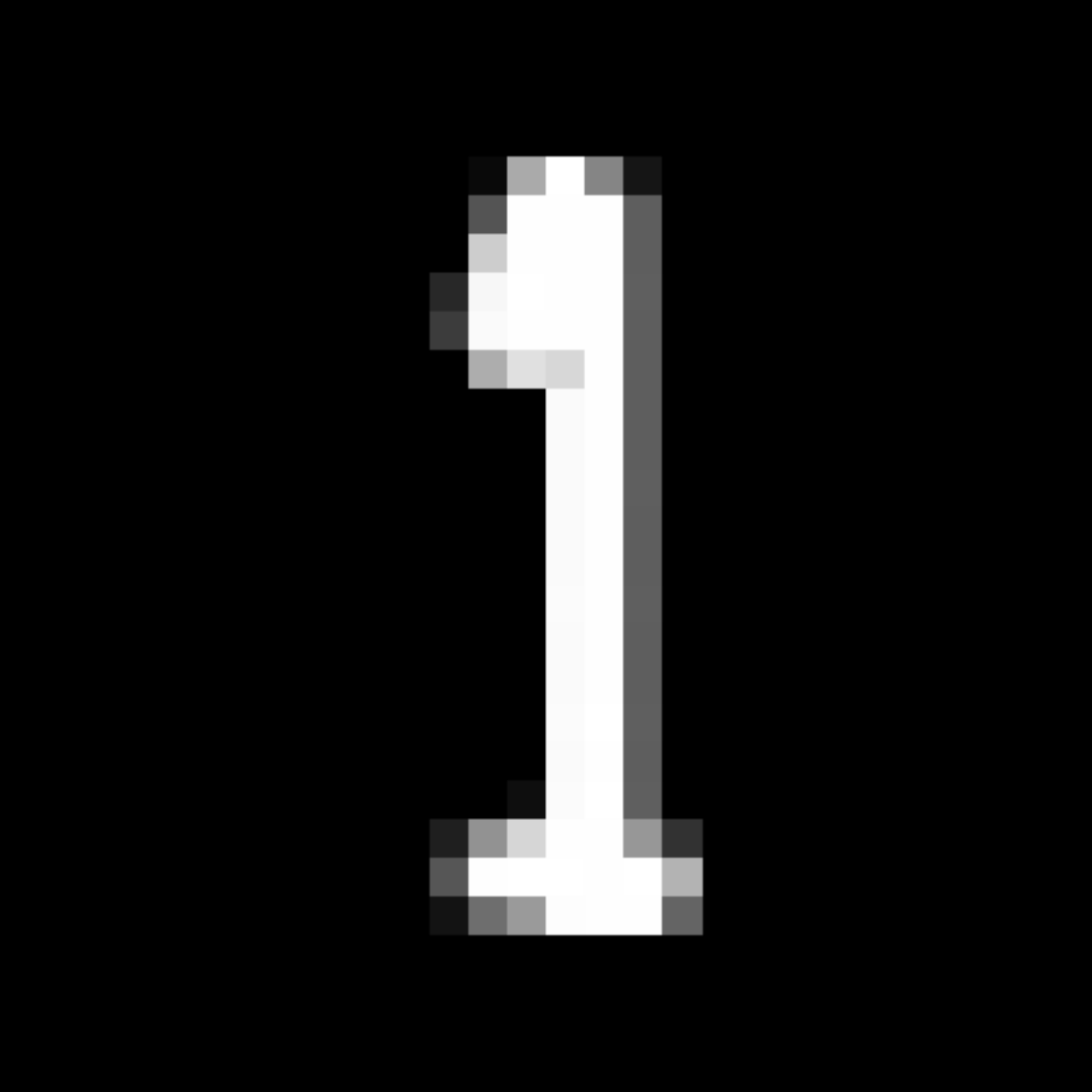}  & \mnist{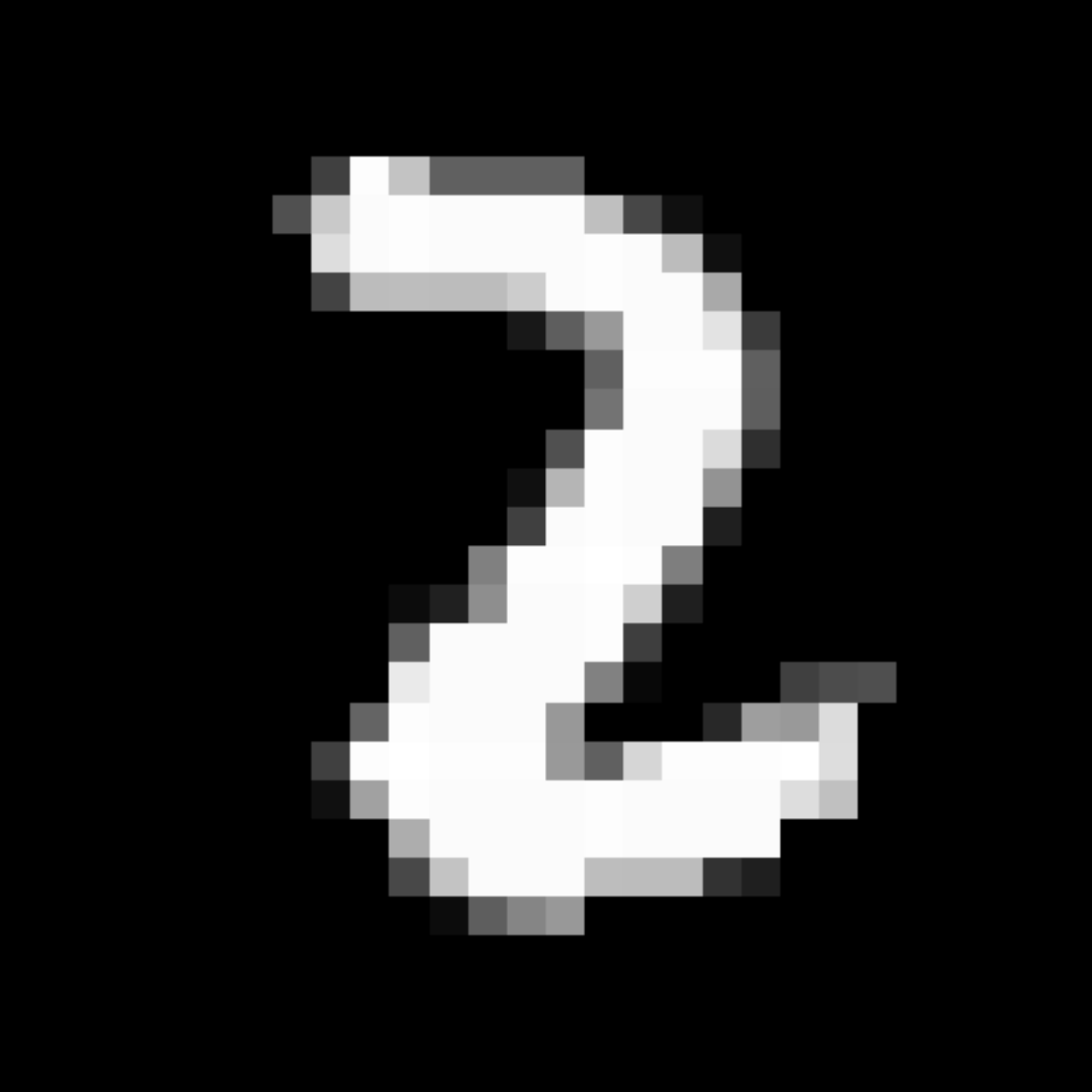}  & \mnist{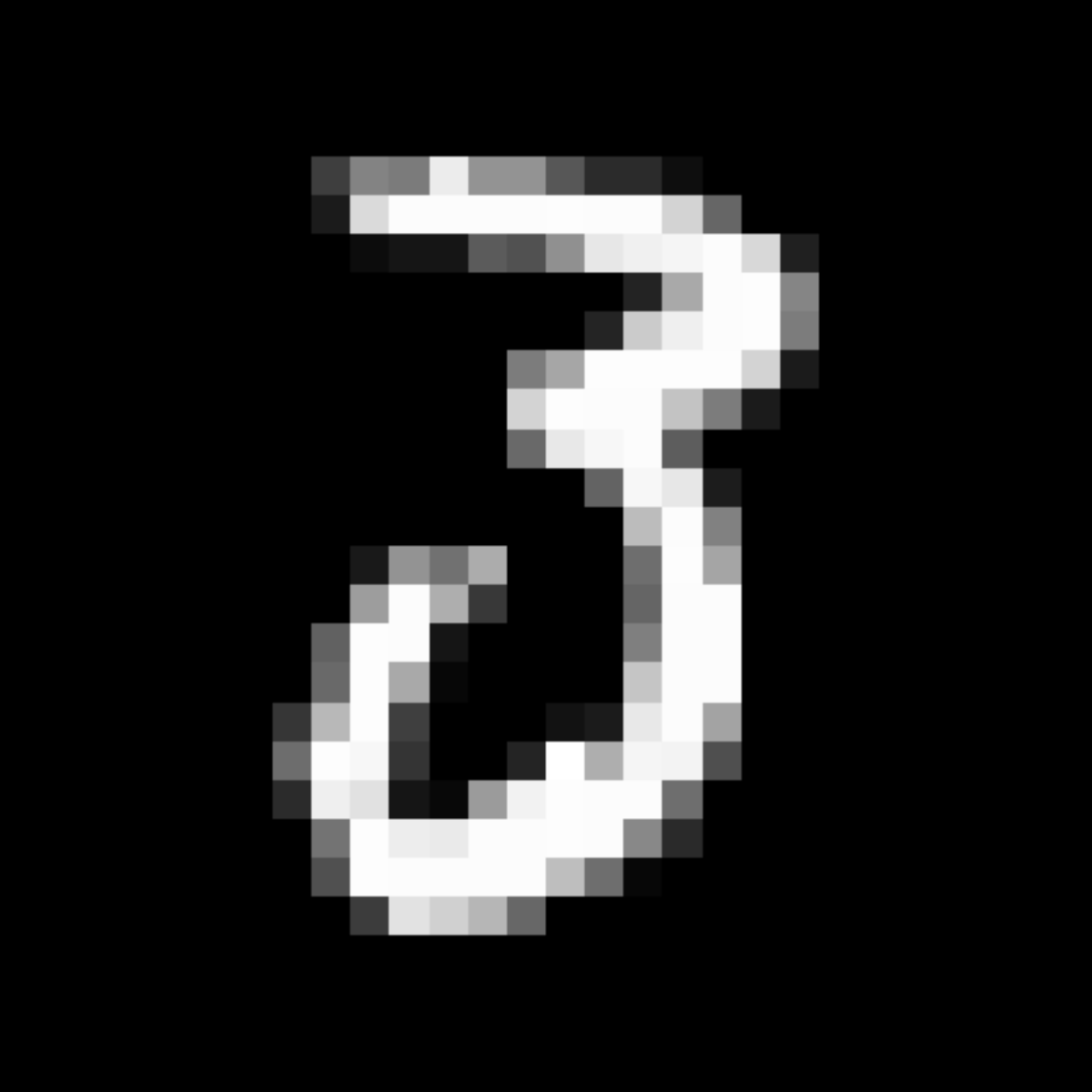}  & \mnist{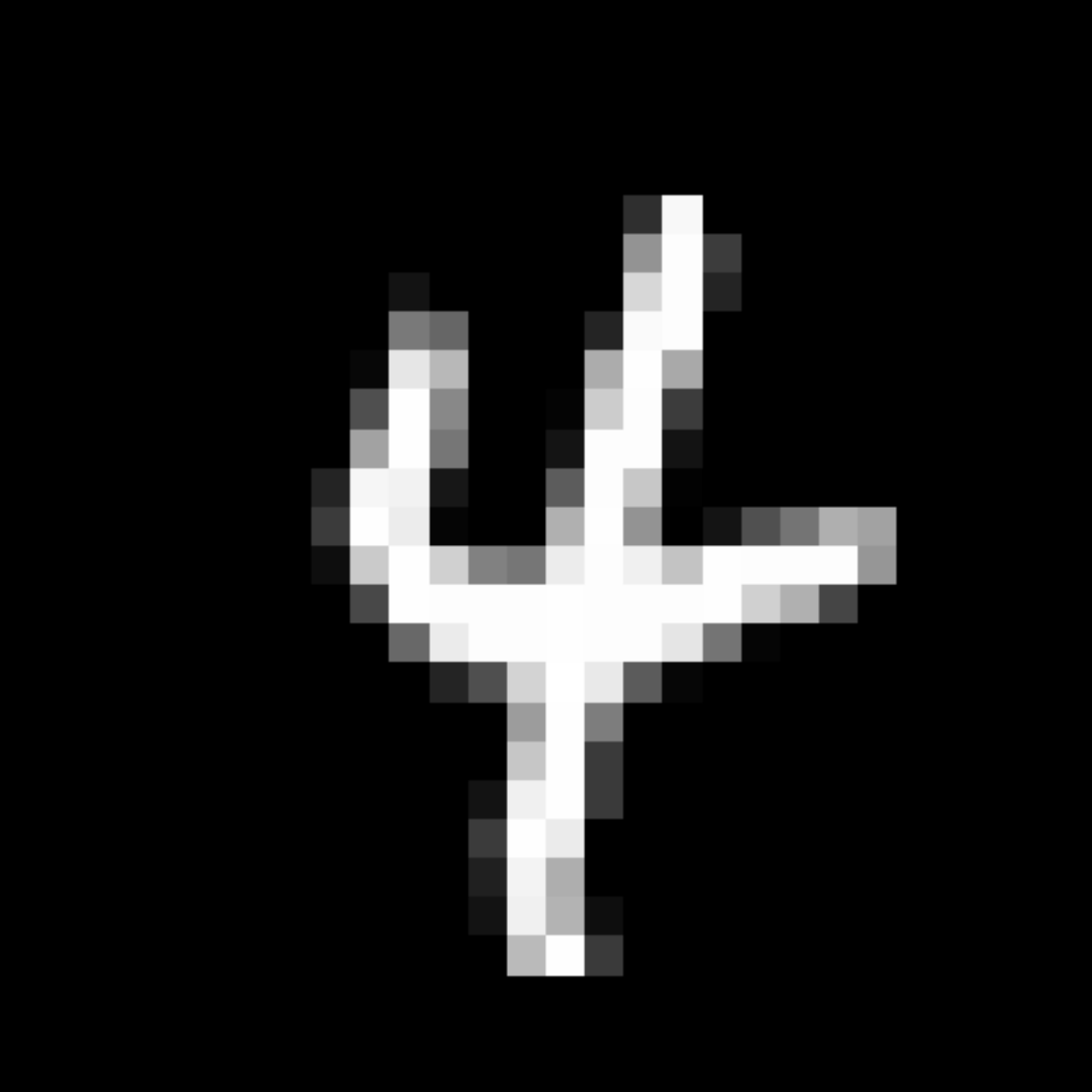}  & \mnist{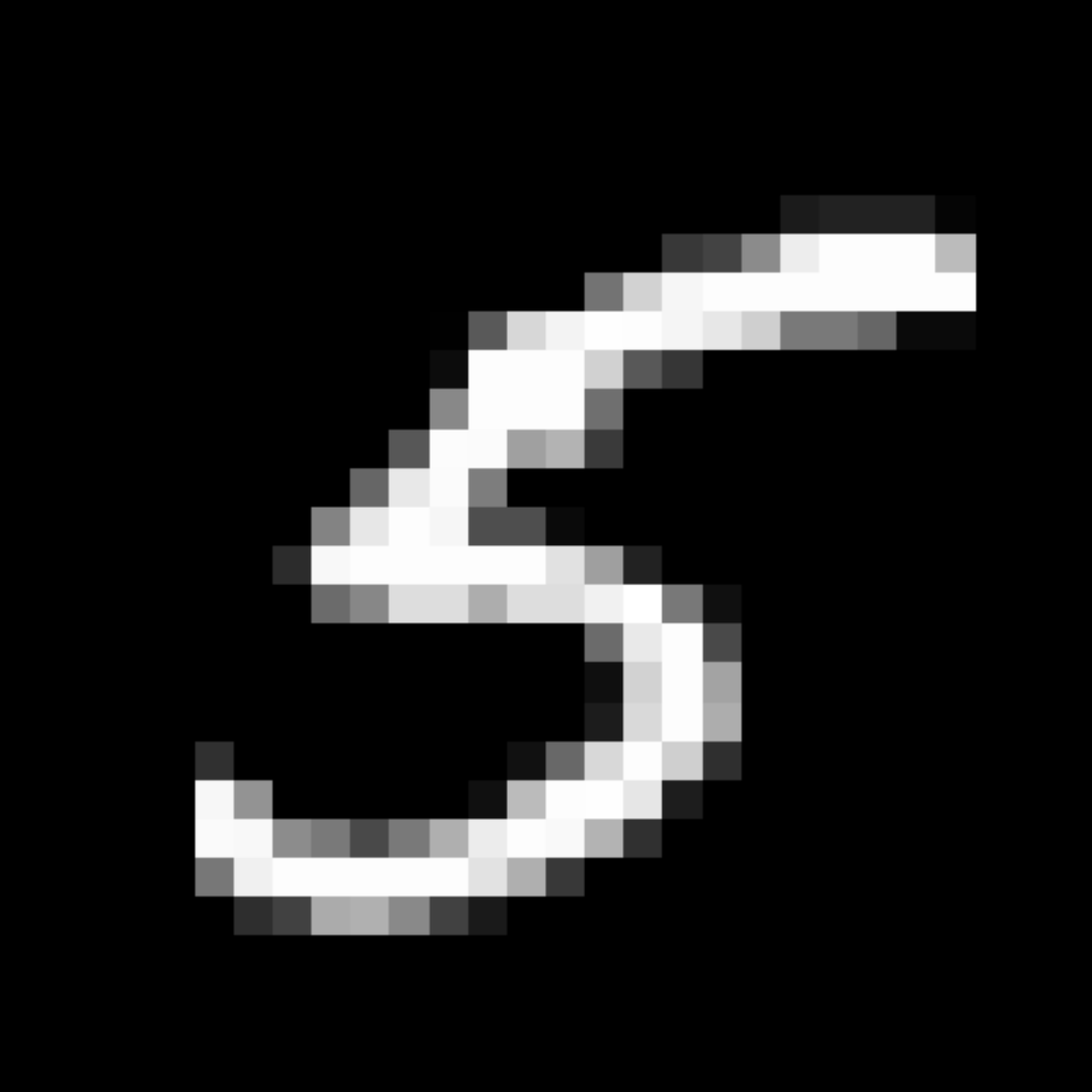}  & \mnist{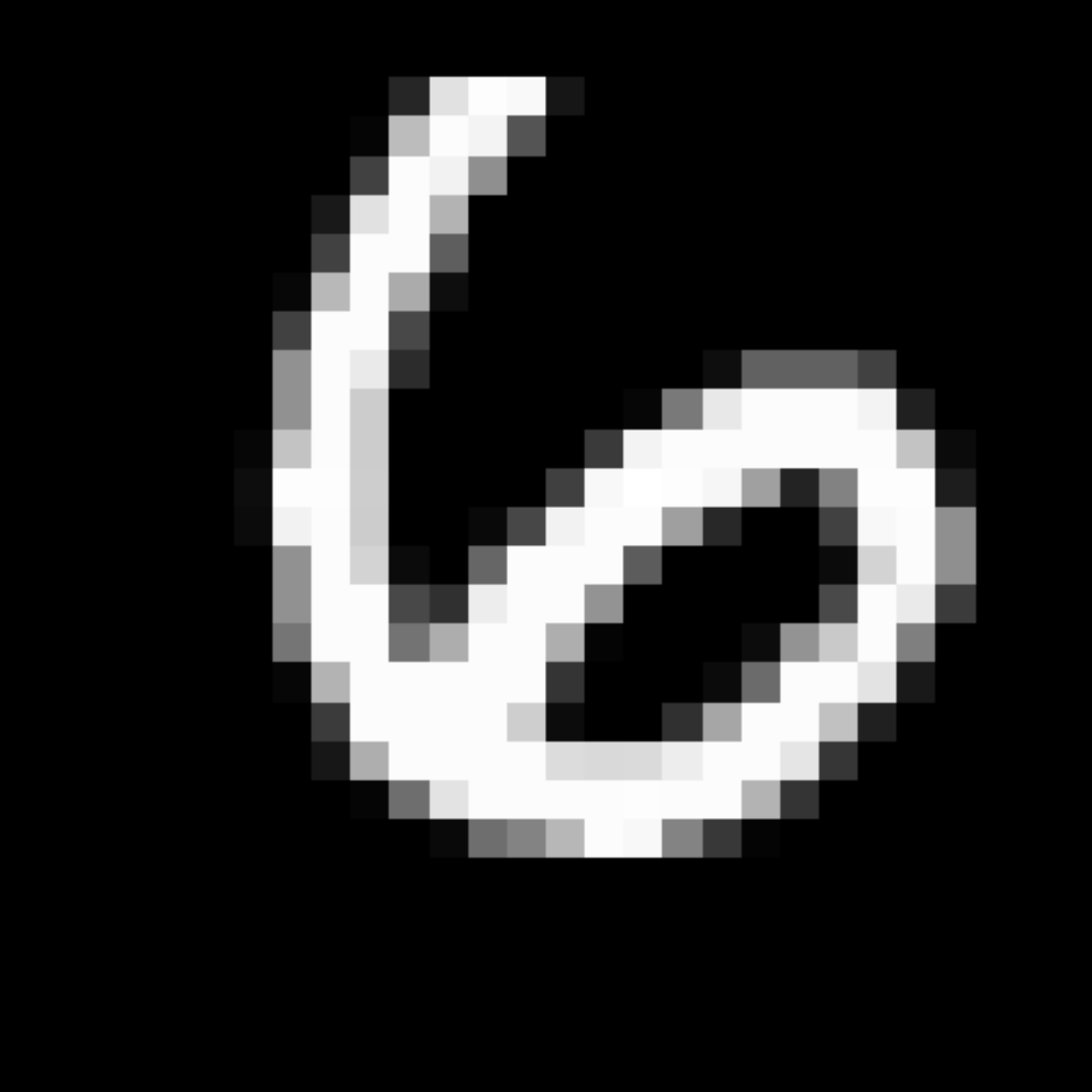}  & \mnist{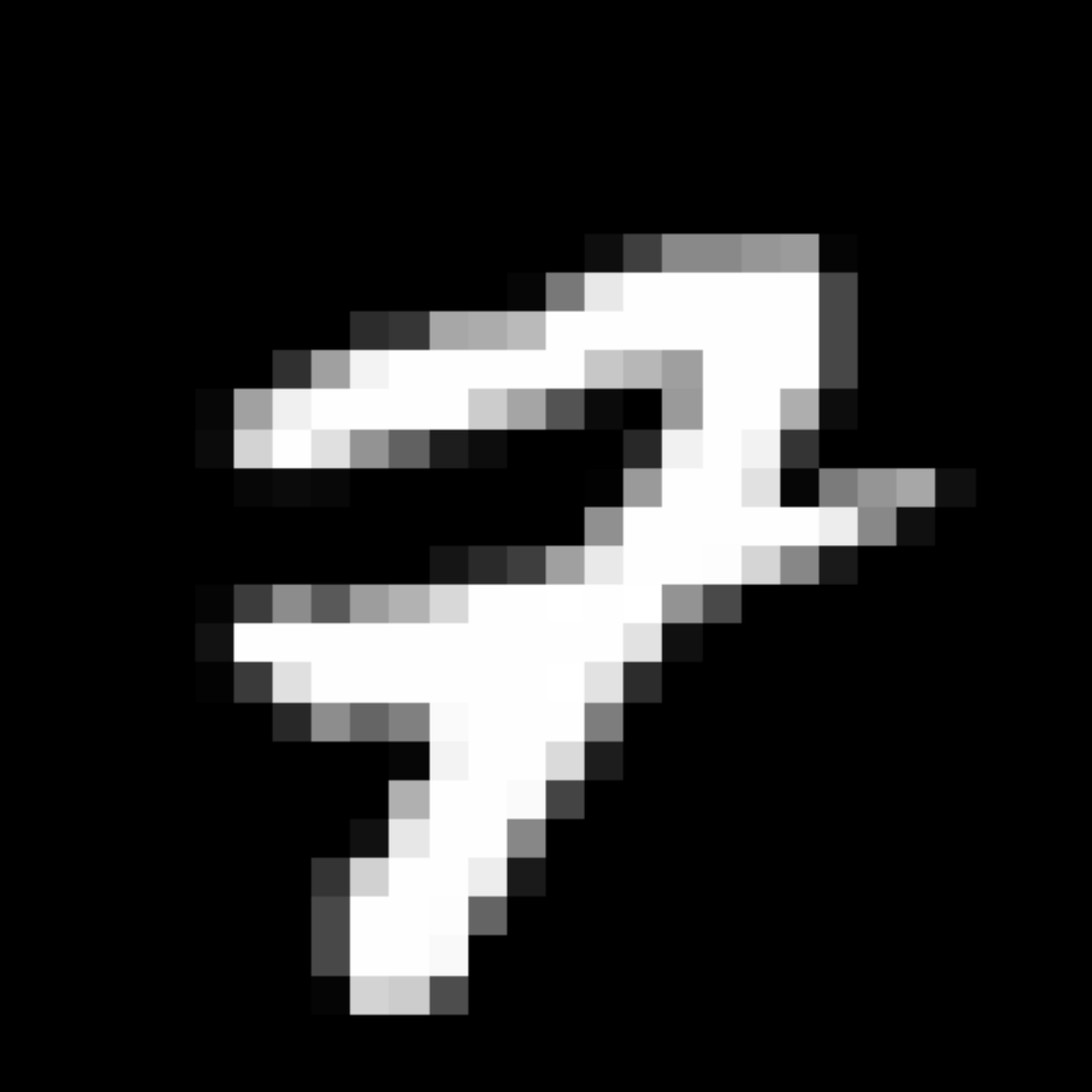}  & \mnist{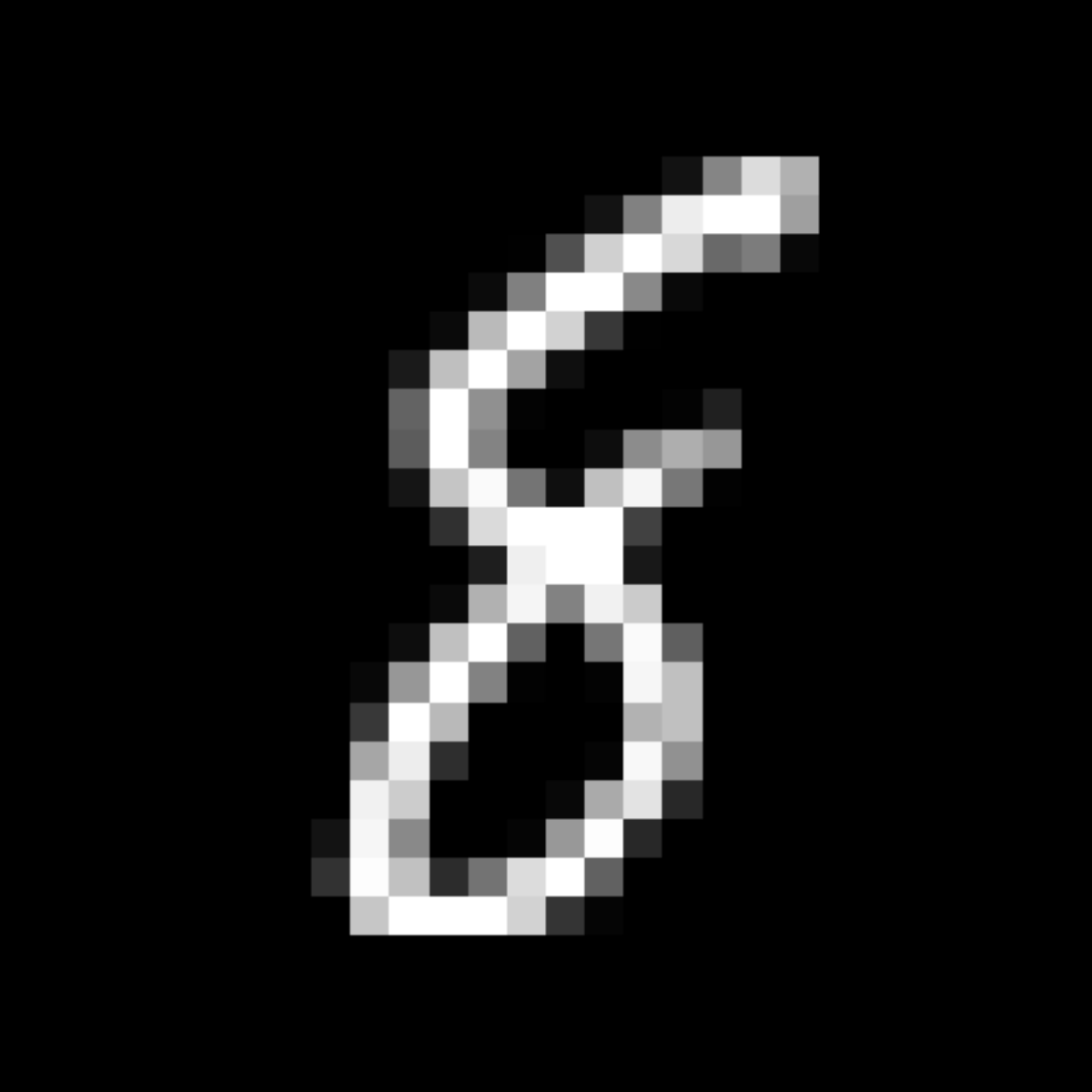}  & \mnist{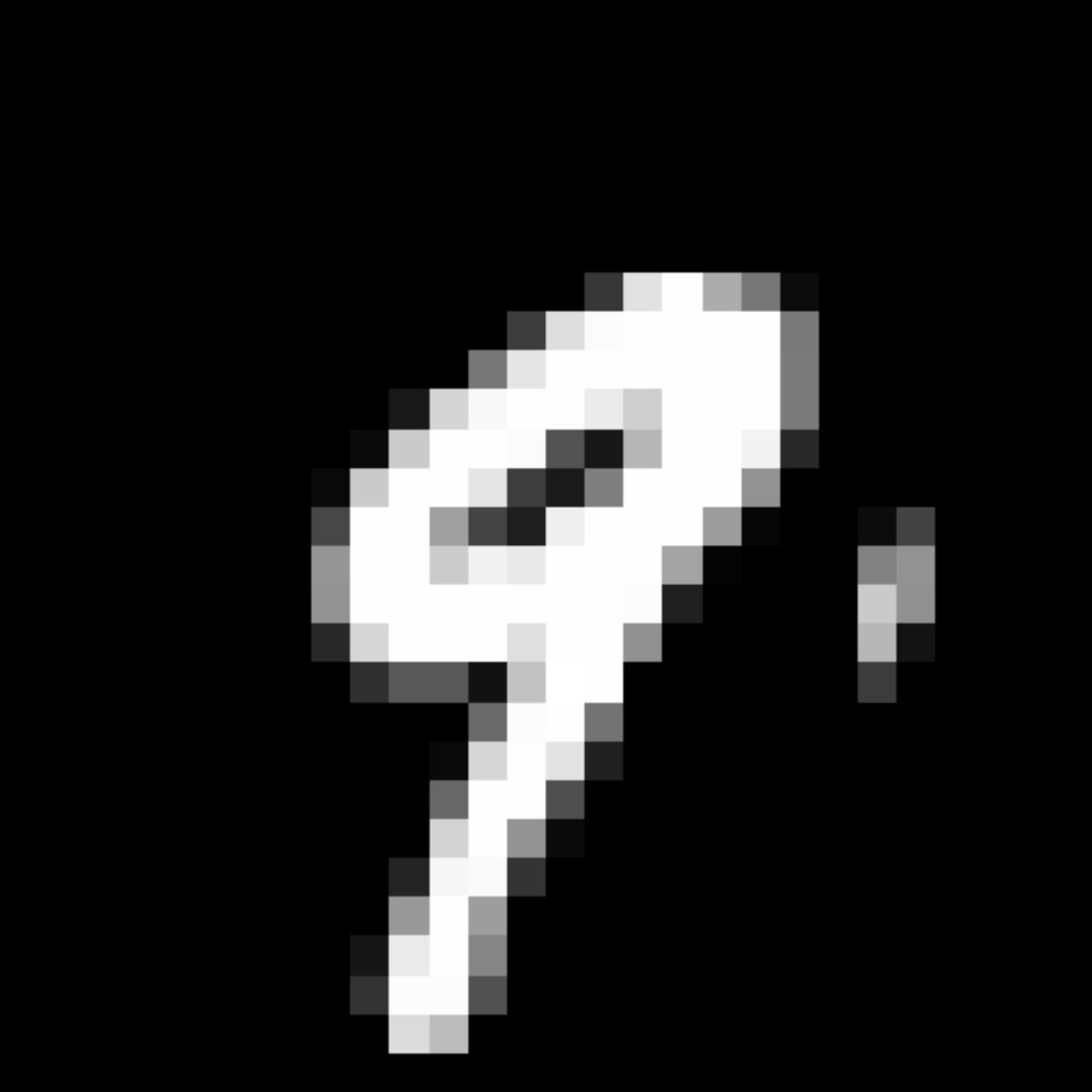}  \\ \hline
		FashionMNIST                           &  \cite{Xiao2017}   & $28$\,$\times$\,$28$ &          60\,000          &        10\,000        & \fmnist{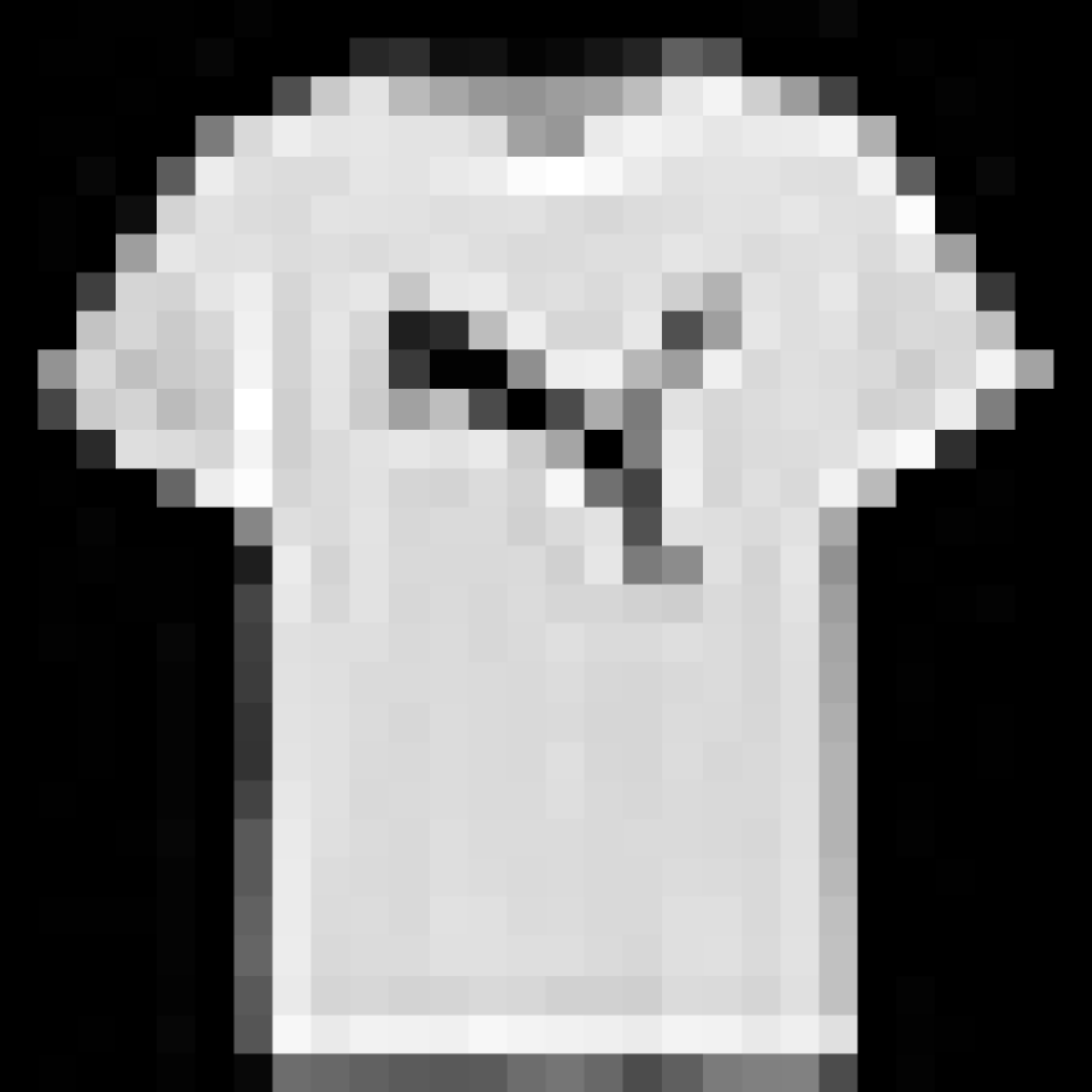} & \fmnist{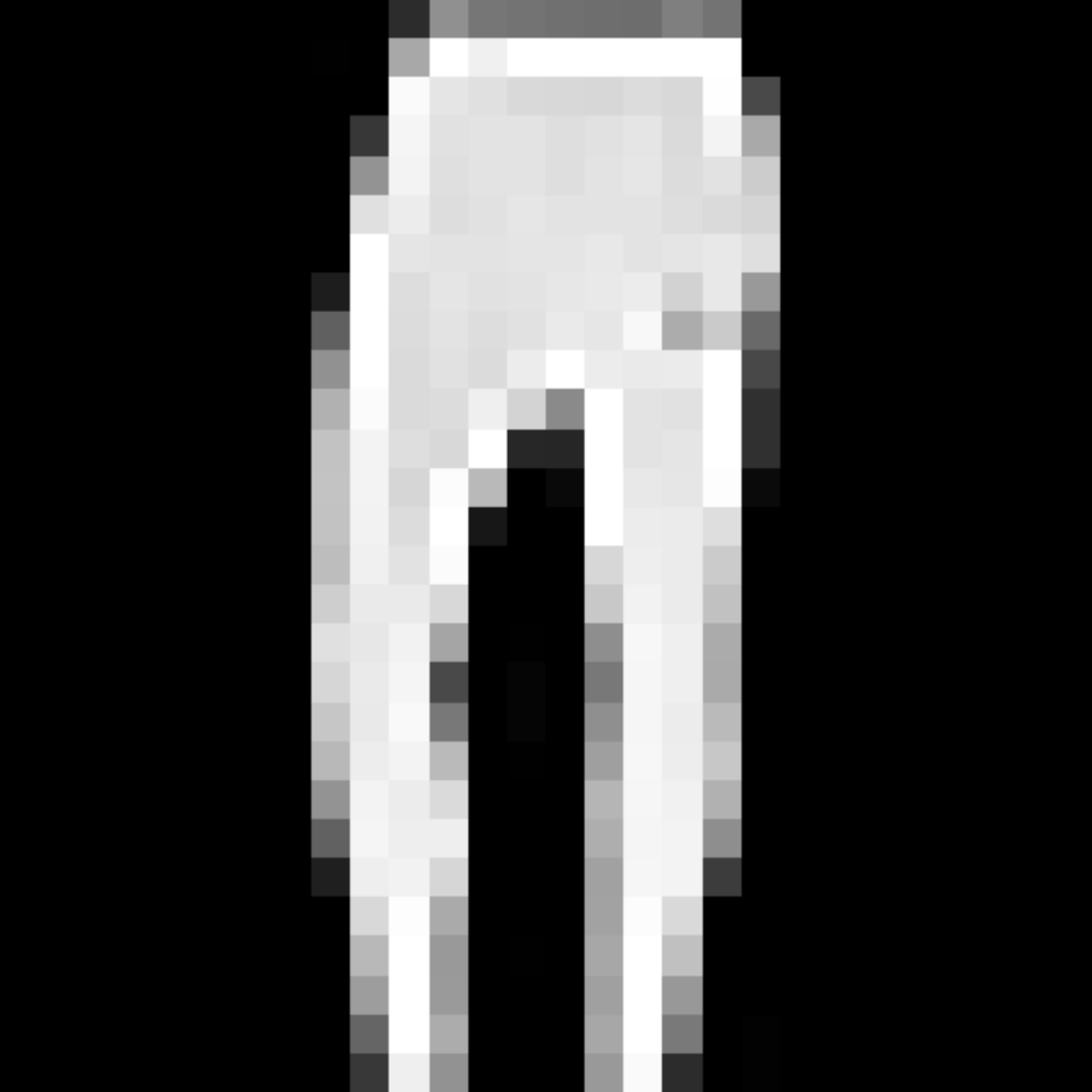} & \fmnist{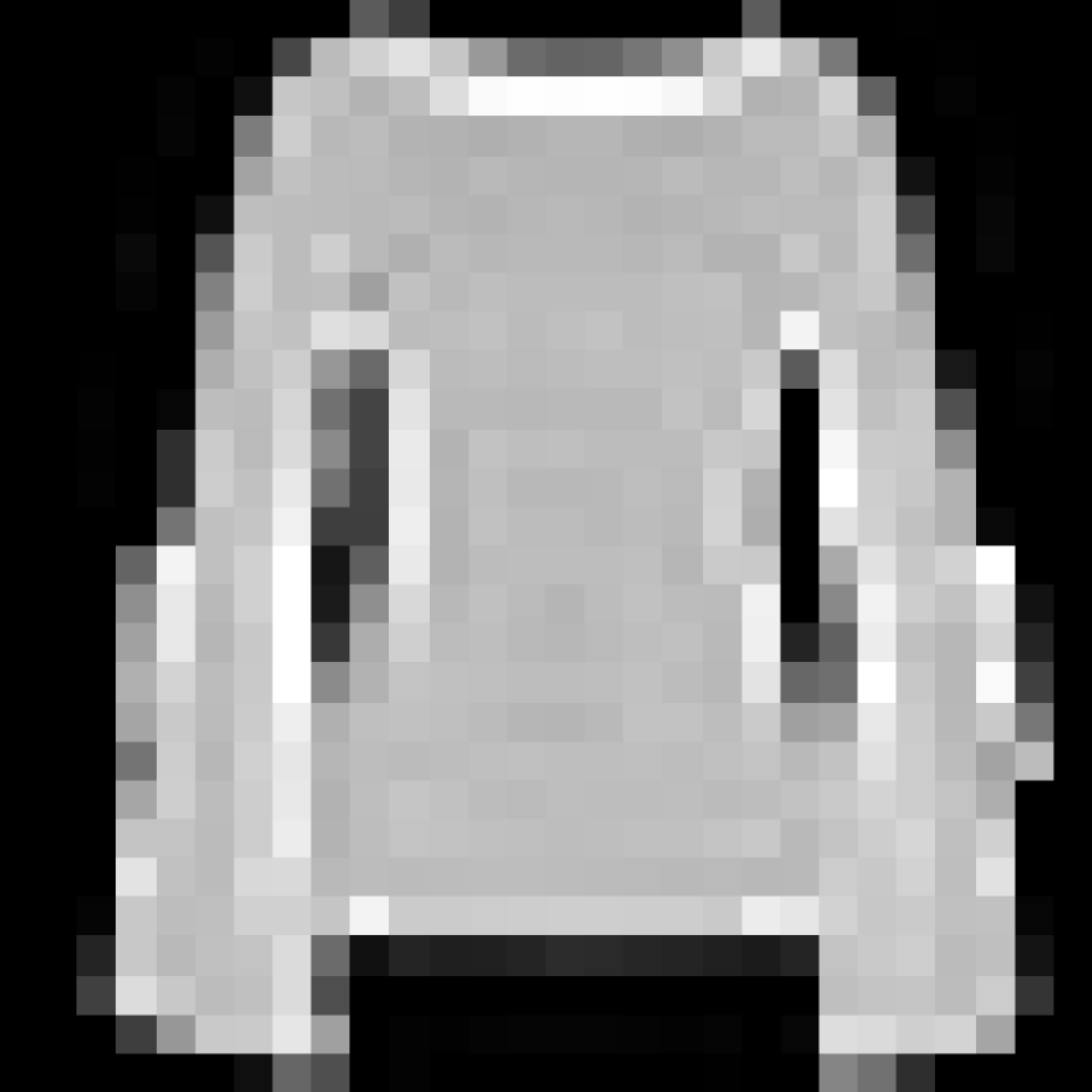} & \fmnist{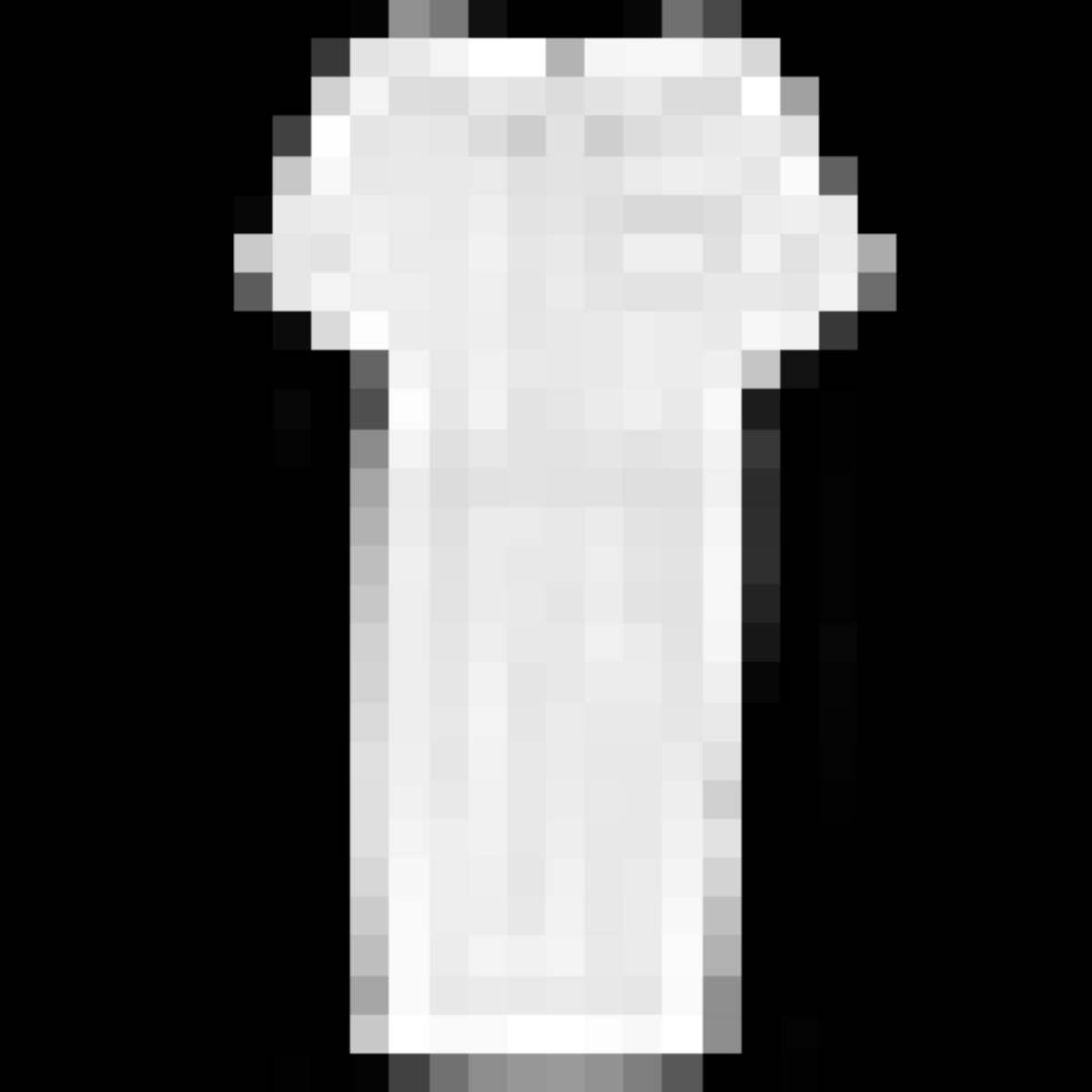} & \fmnist{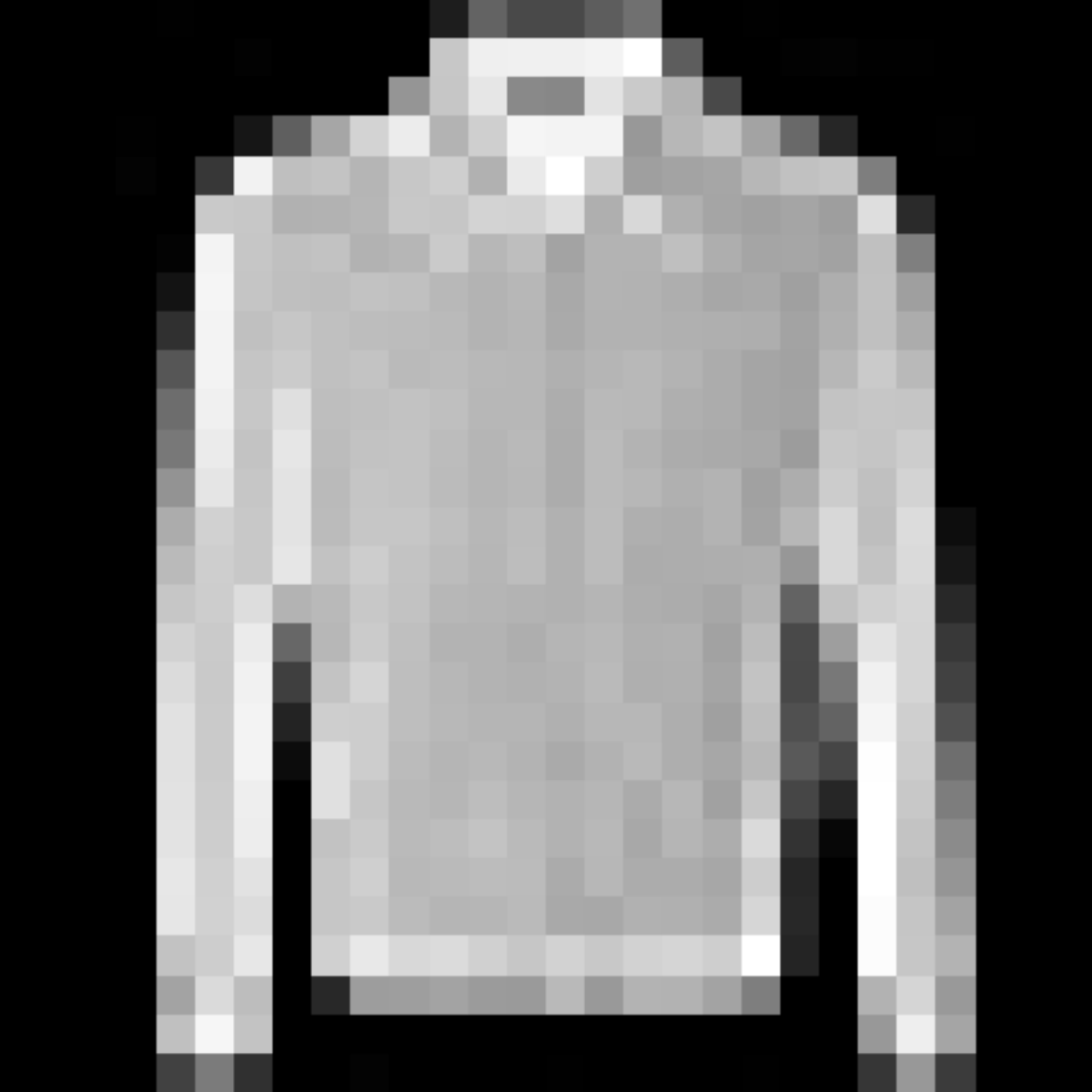} & \fmnist{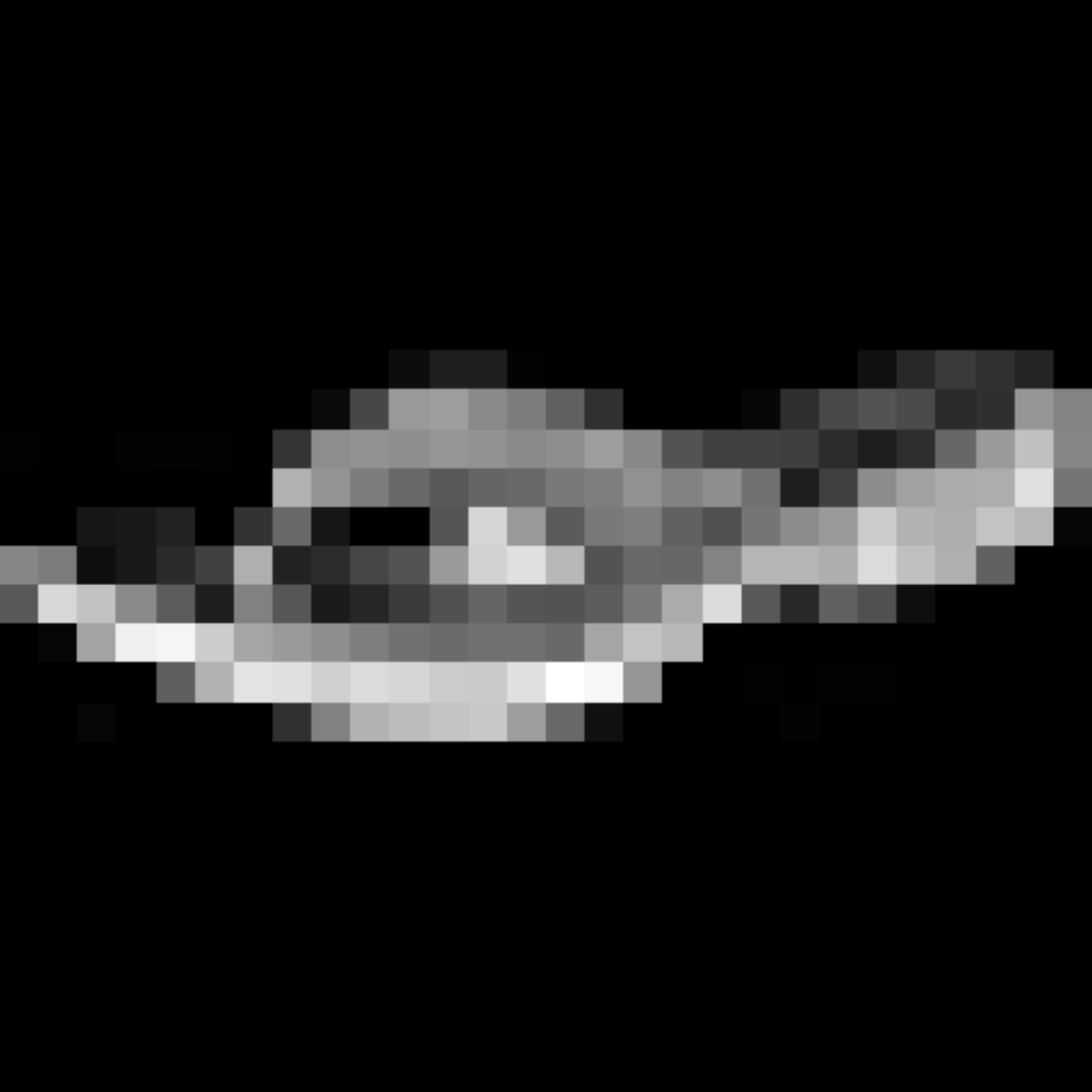} & \fmnist{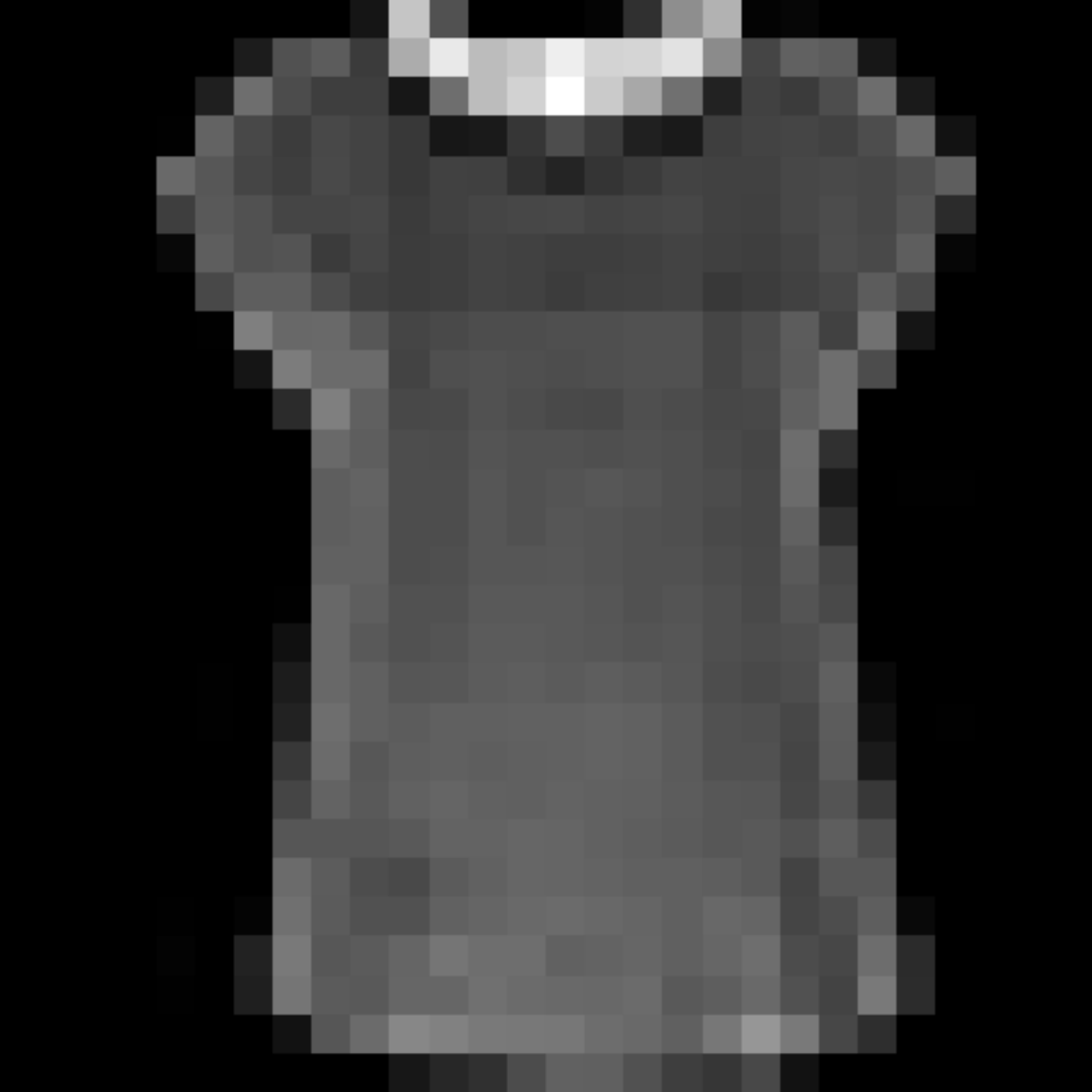} & \fmnist{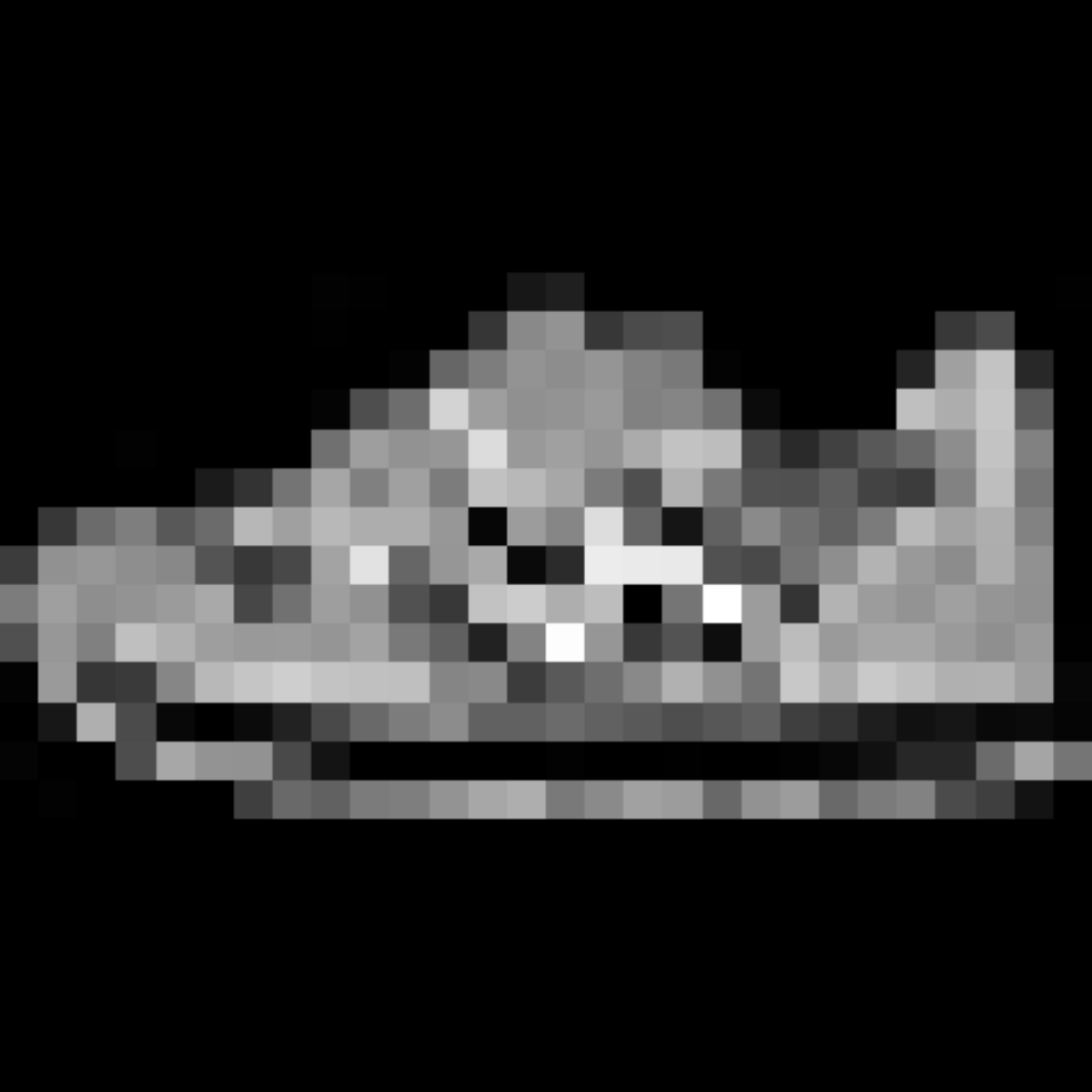} & \fmnist{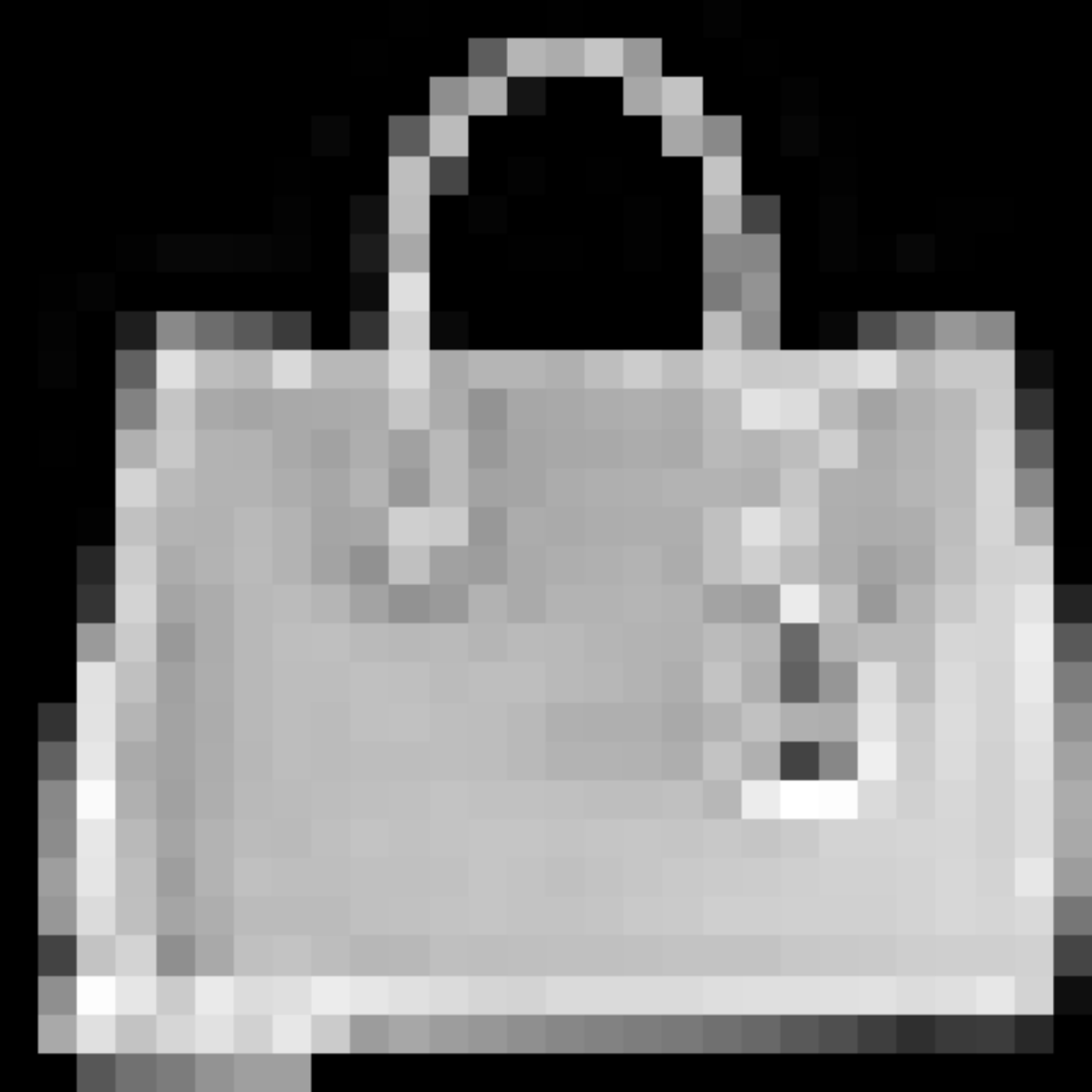} & \fmnist{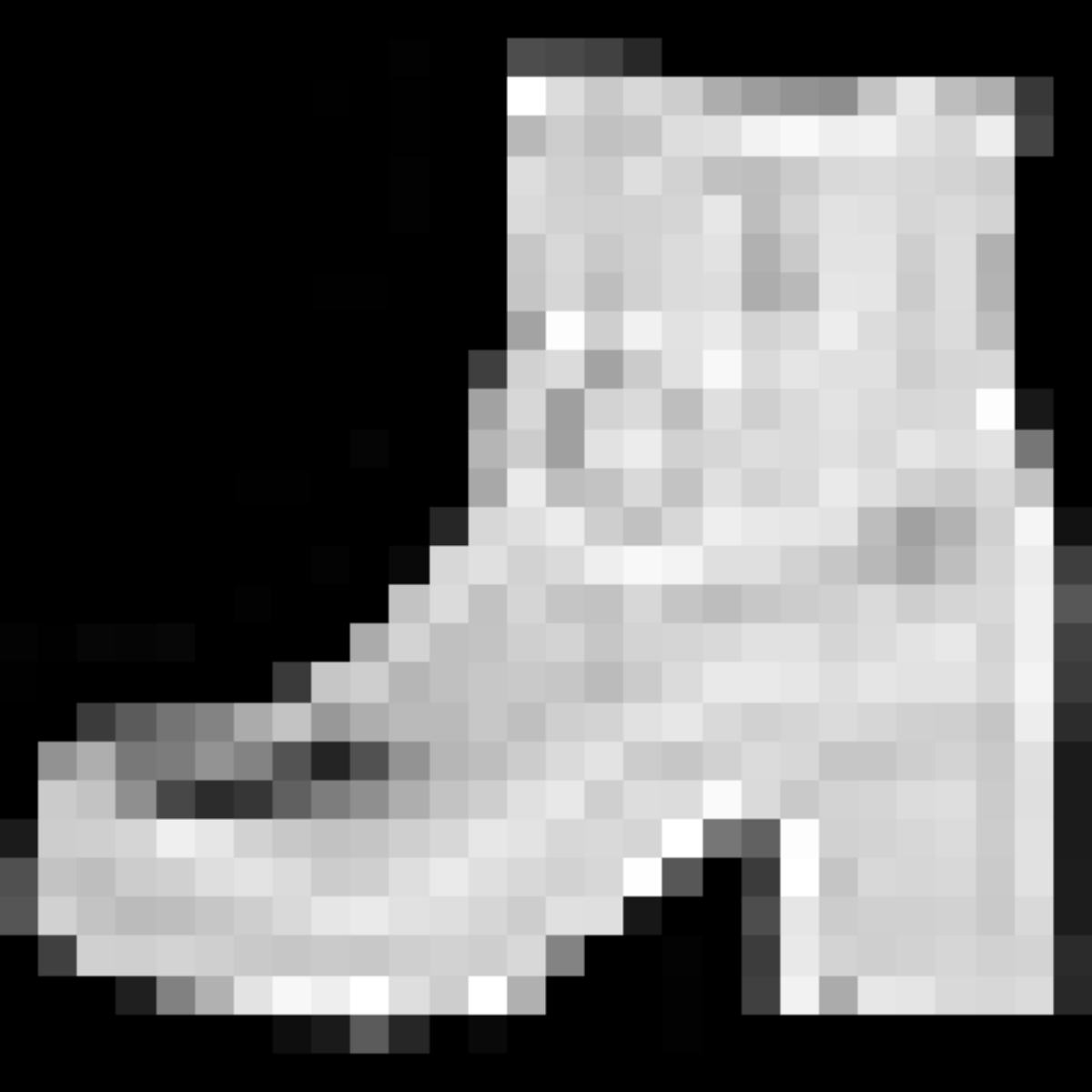} \\ \hline
		Devanagari                             & \cite{Acharya2015} & $32$\,$\times$\,$32$ &          18\,000          &   \phantom{1}2\,000   & \devan{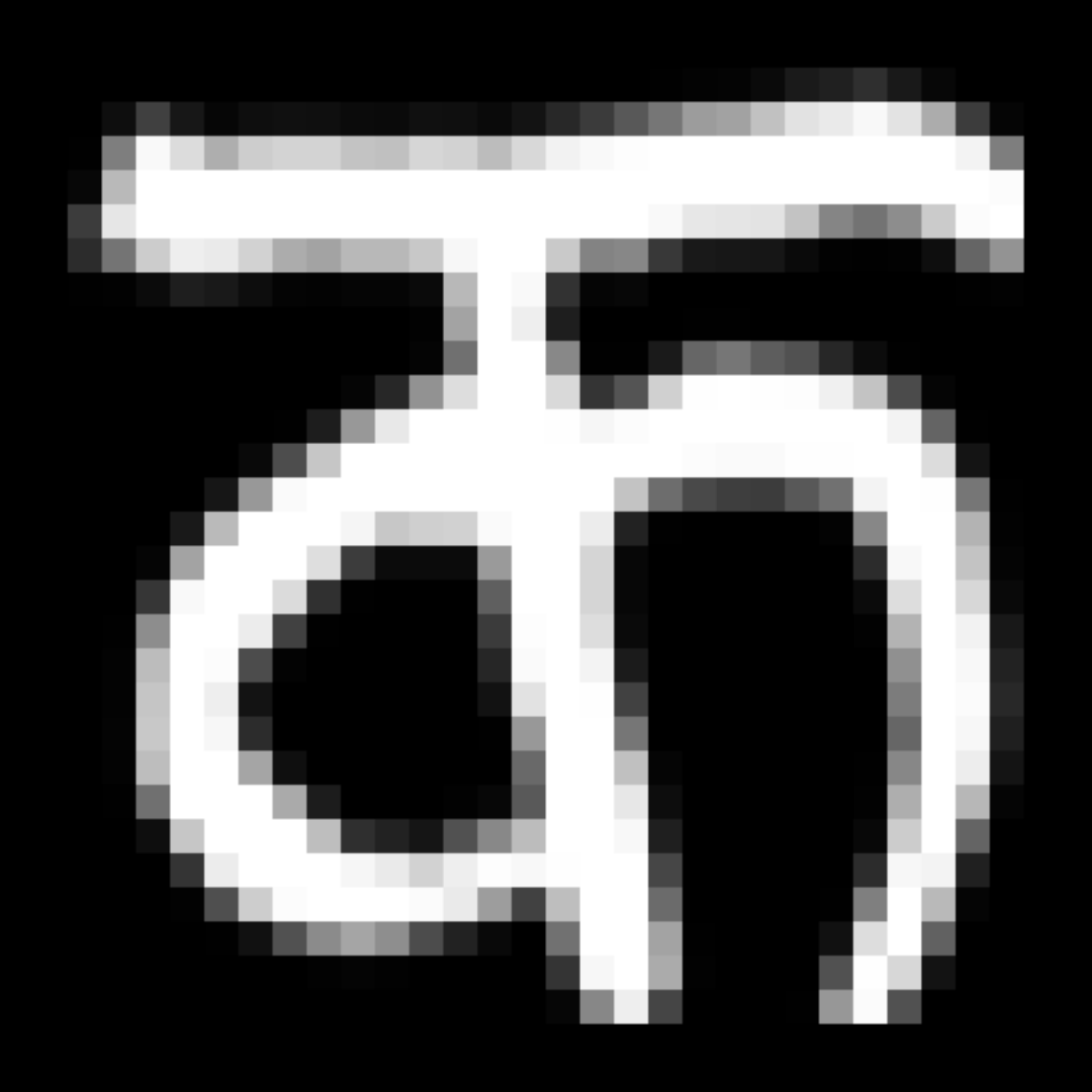}  & \devan{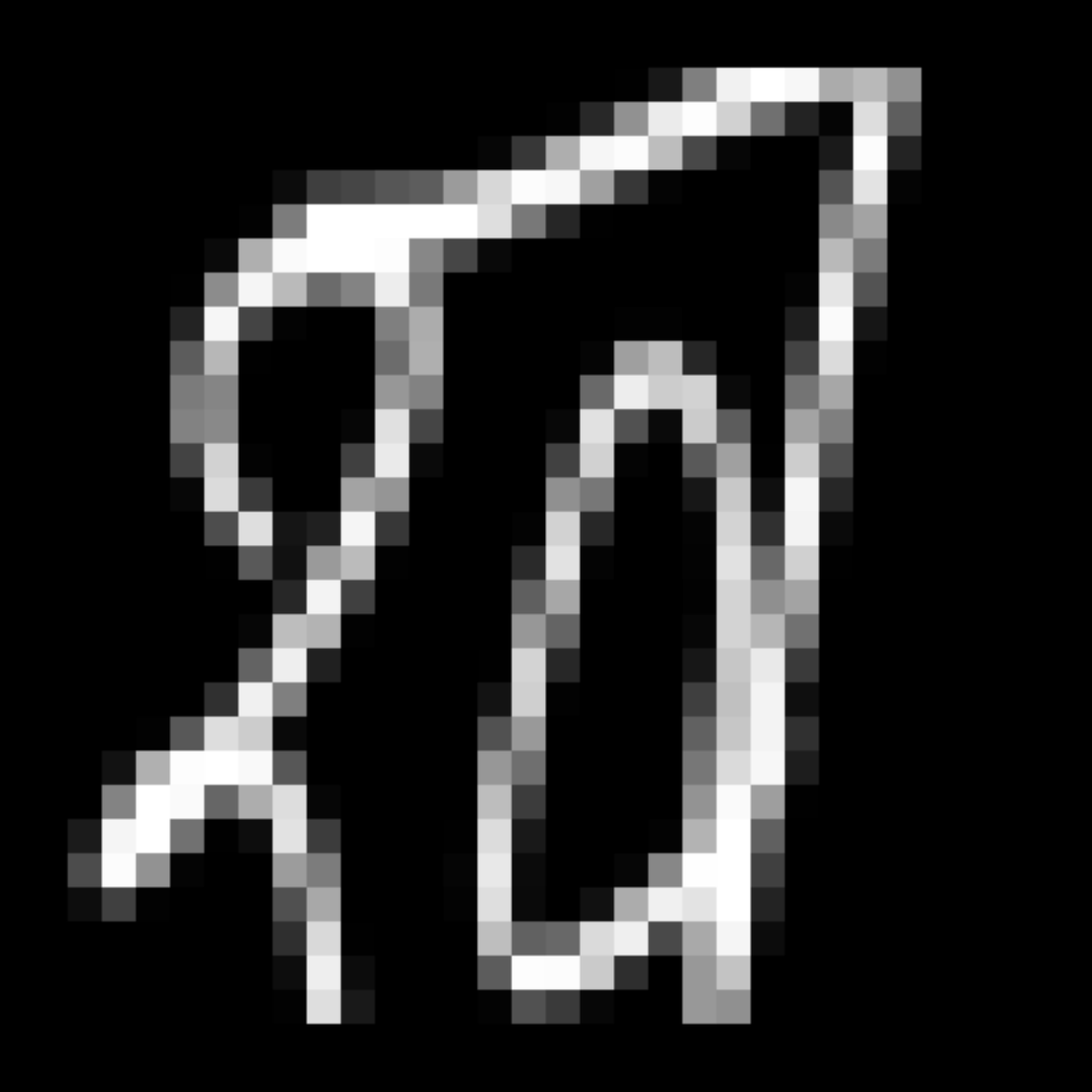}  & \devan{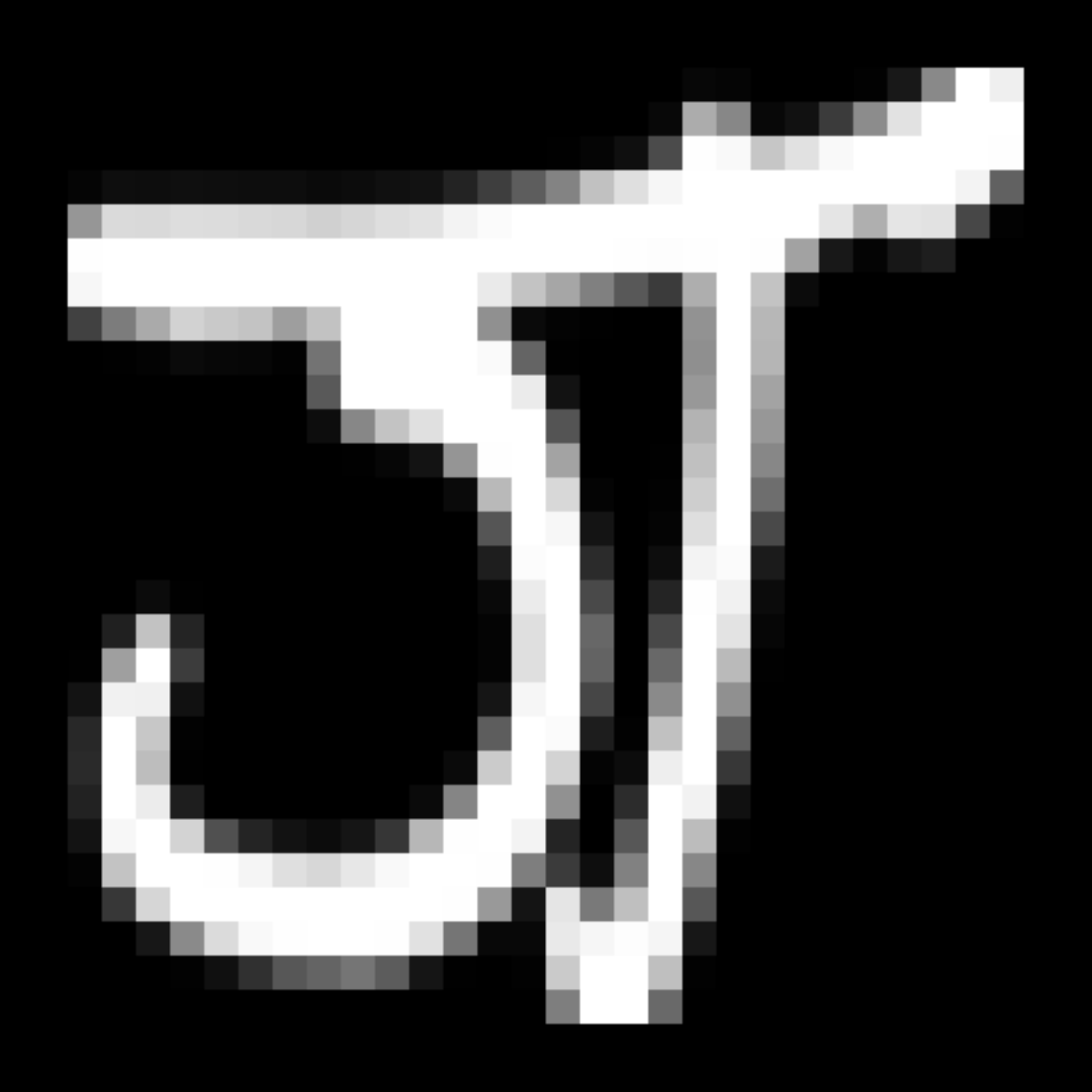}  & \devan{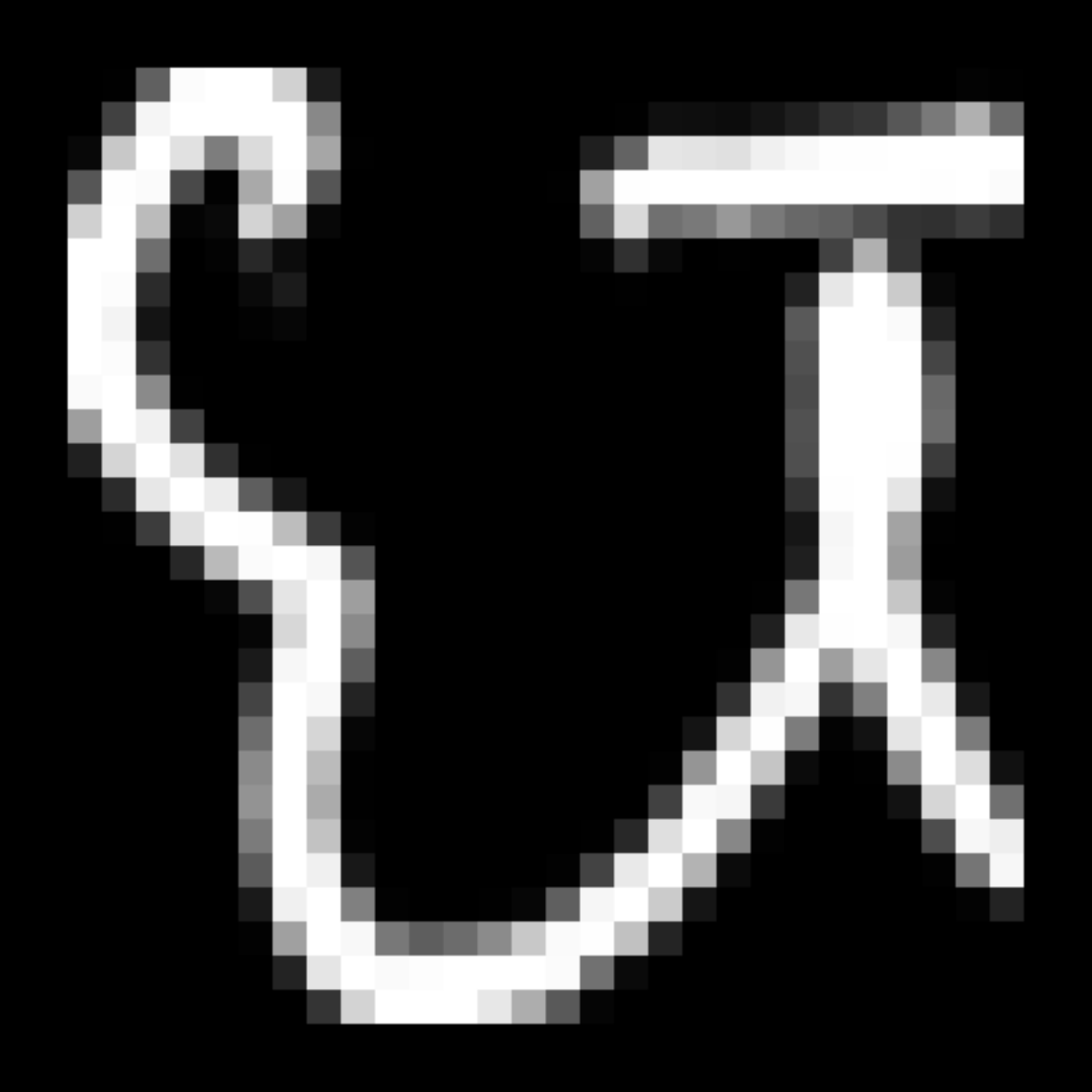}  & \devan{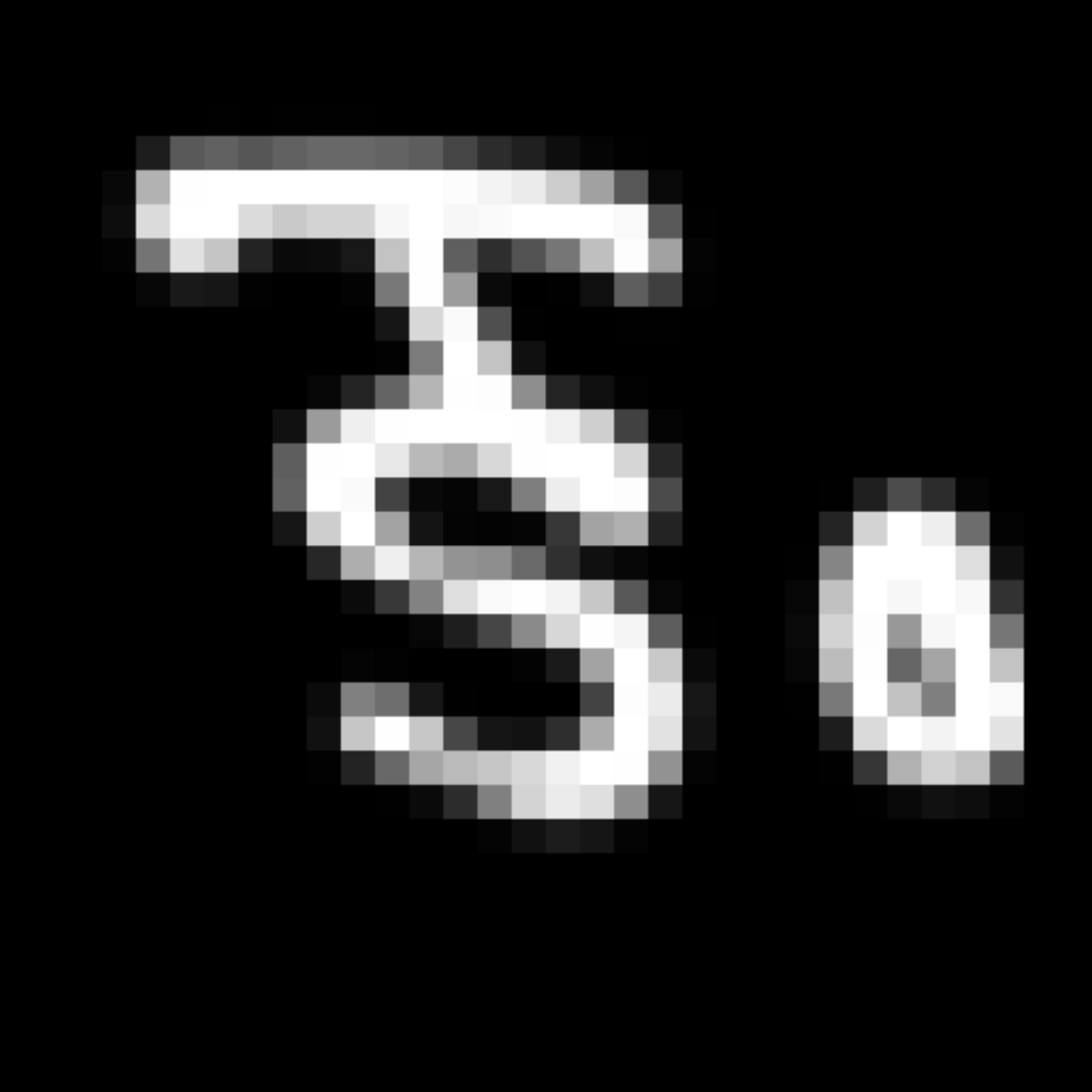}  & \devan{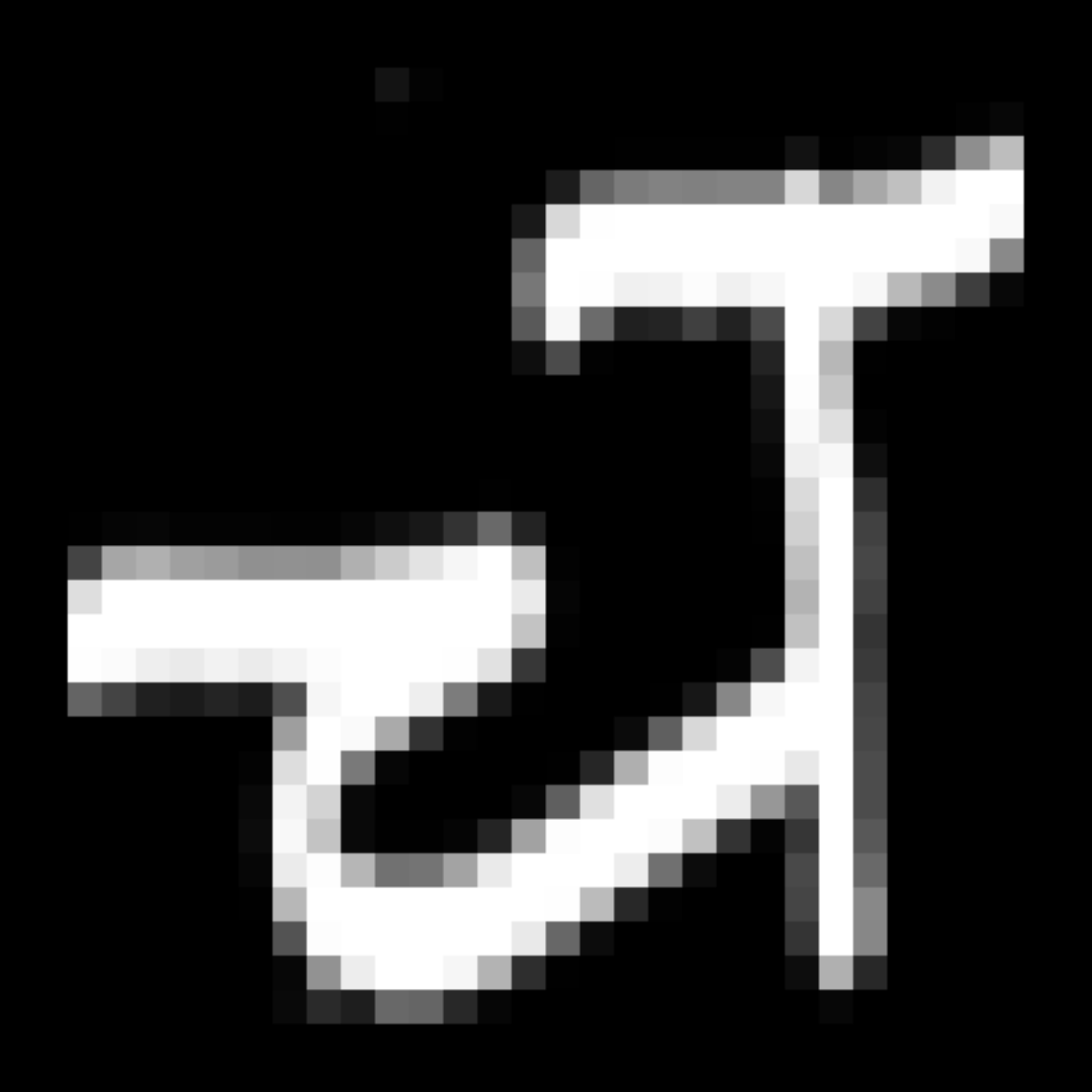}  & \devan{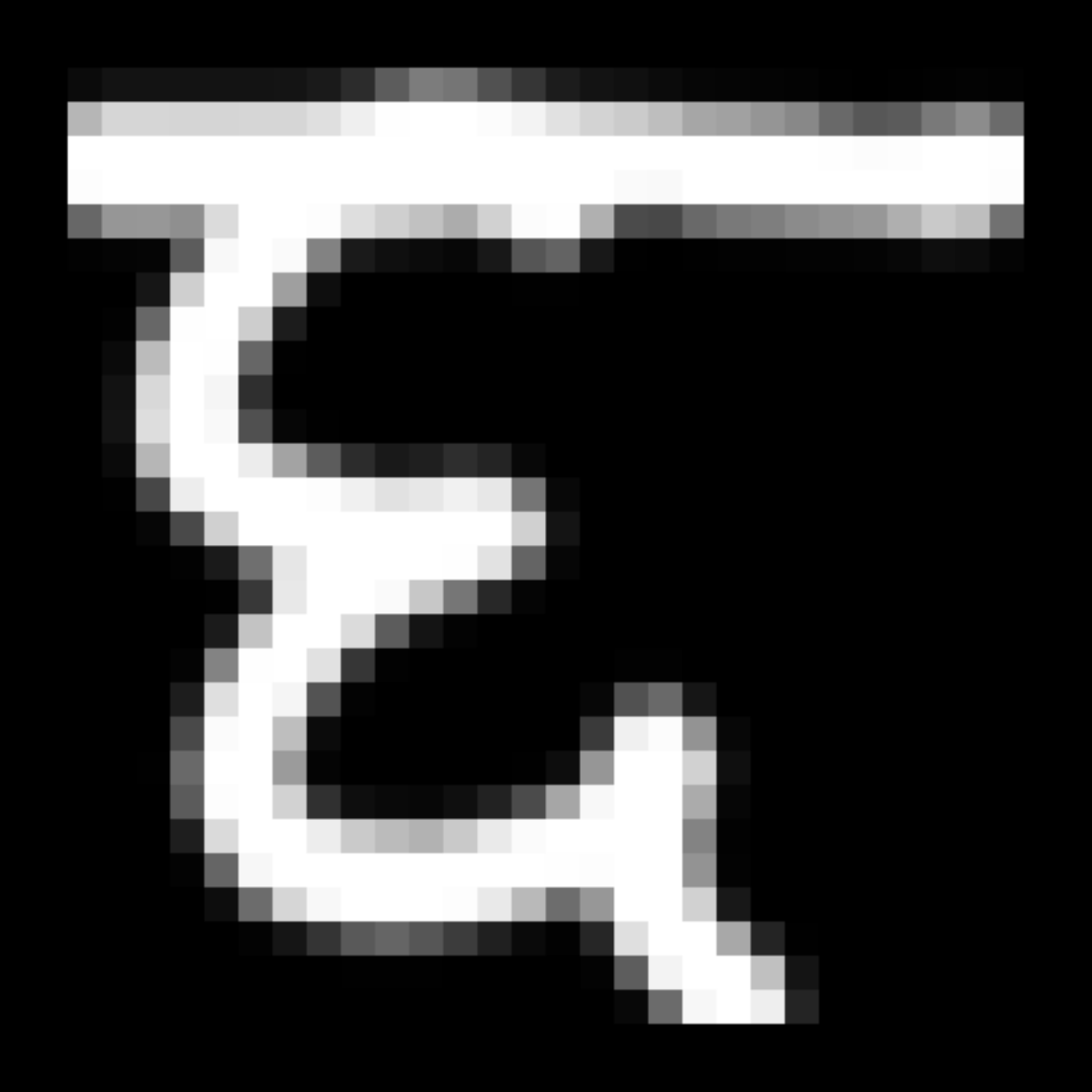}  & \devan{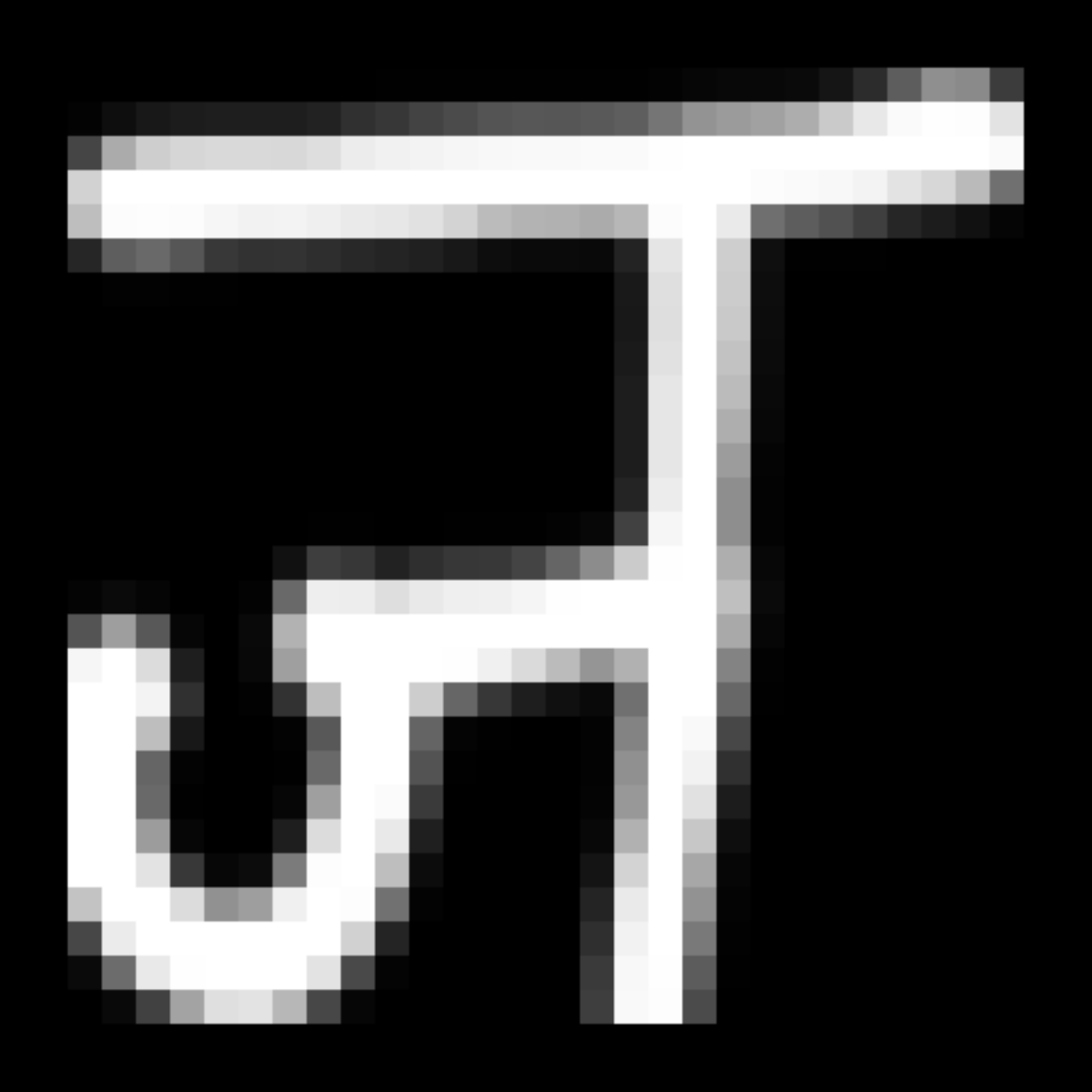}  & \devan{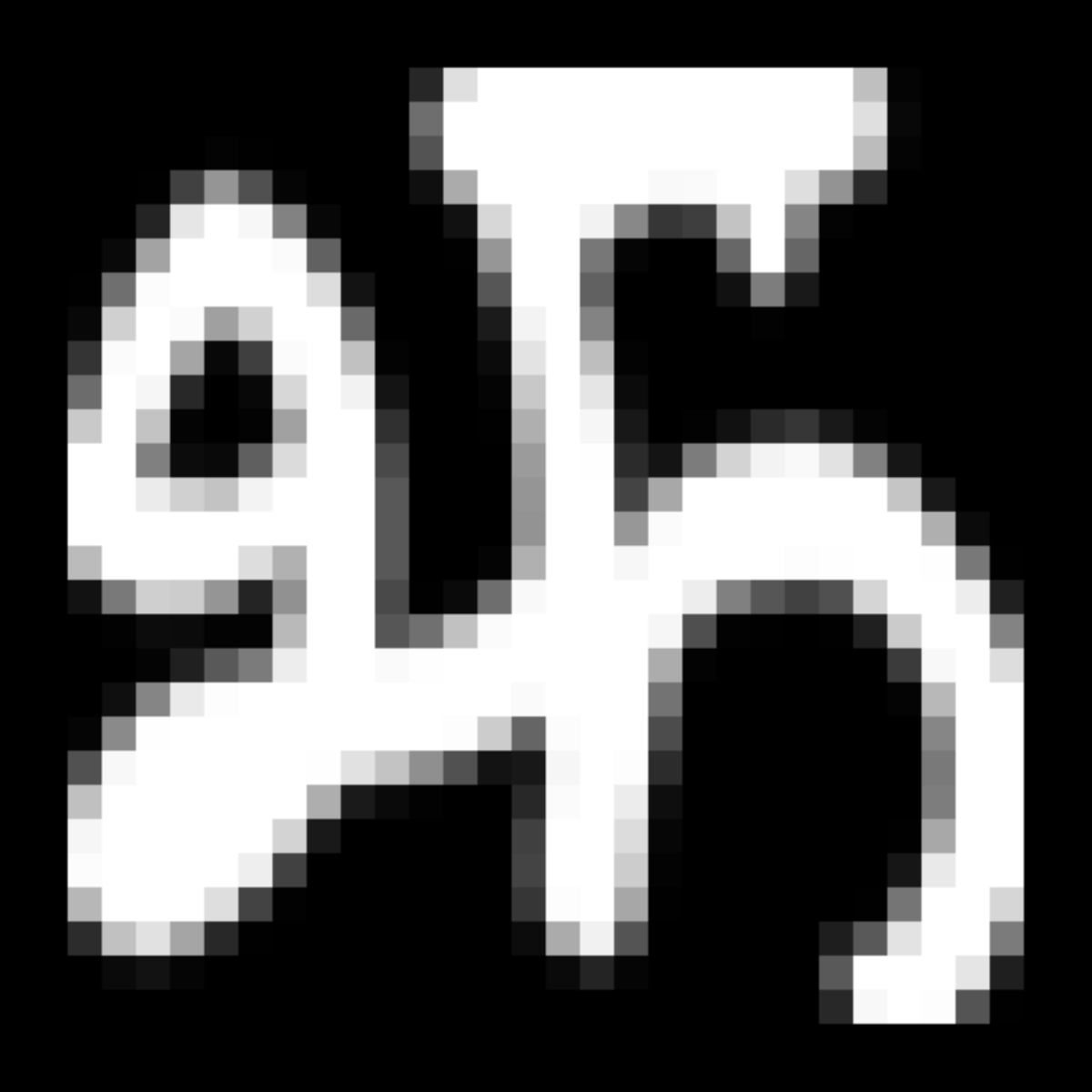}  & \devan{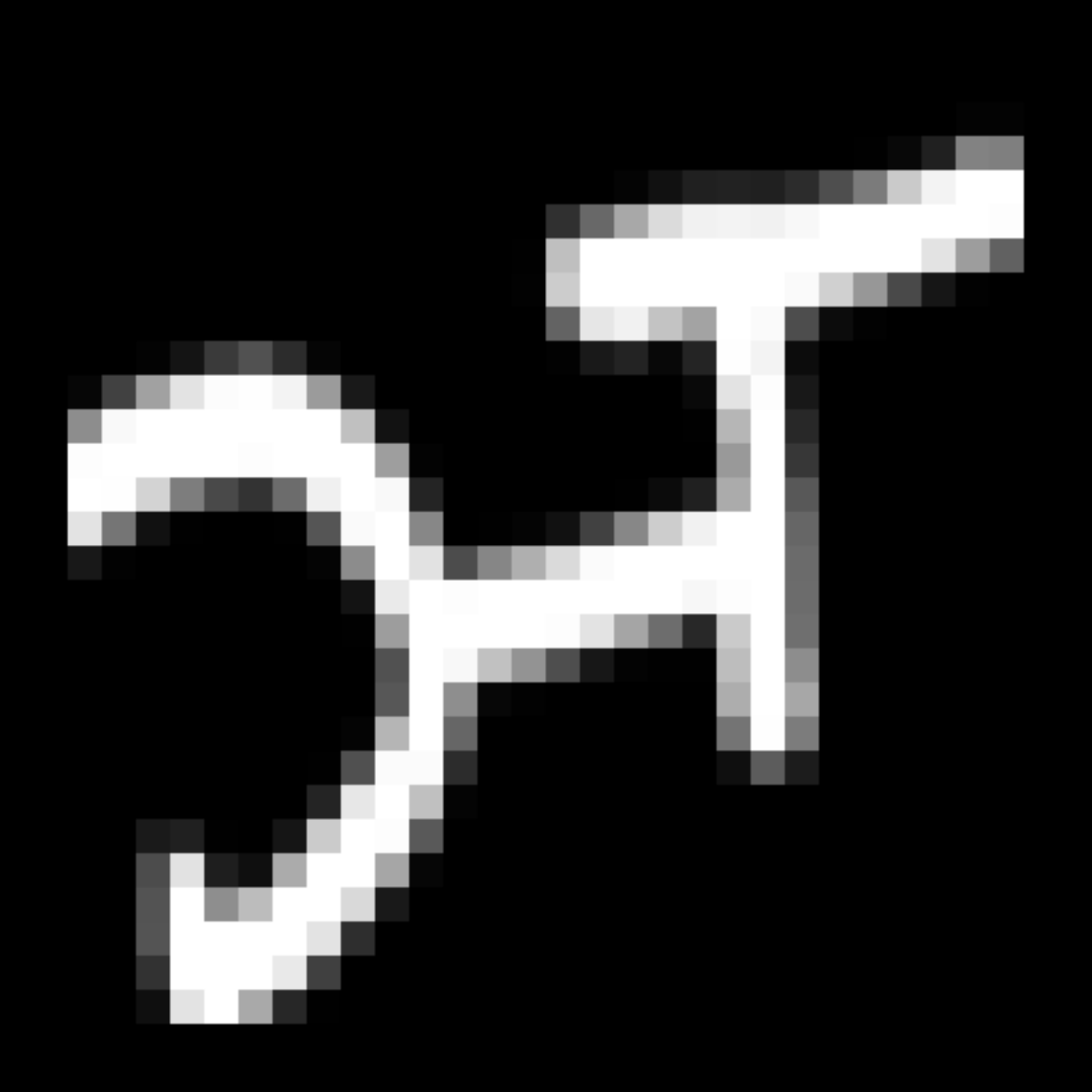}  \\ \hline
	\end{tabular}
\end{table*}
\section{Gaussian Mixture Replay in Detail}\label{sec:gmr-details}
\noindent As stated in \cref{sec:gmr_intro}, GMR is comprised of a generator realized by a GMM, and a learner realized by a linear classifier. 
Both can indeed be replaced by more complex, \enquote{deeper} methods, but we limit our work to simplest case.F
\par
The generator consists of $K$ Gaussian mixture components, each maintaining a separate $\vmu_k$ centroid and covariance matrix $\mSigma_j$. 
Covariance matrices are always taken to be diagonal (a justification for this is given in the discussion). 
As the basic data flow in GMR has been outlined in \cref{sec:gmr_intro}, we will describe the procedure for training, sampling and outlier detection along with the principal GMR hyper-parameters.
\subsection{Outlier Detection}\label{sec:gmr-outliers}
\noindent Outlier detection is performed by the generator according to standard GMM procedures. 
Essentially, it is based on the long term average value ($\mathbb{E}$) of the loss function for a given sample.
Anything too far below the \enquote{normal} loss value is considered an outlier. 
To achieve this, we compute the mean and the variance of the GMM loss during training:
\begin{equation*}
	\begin{split}
		\hat\mu(\mathcal L)      & = \mathbb E_i \mathcal L(\vx_i)                        \\
		\hat\Sigma^2(\mathcal L) & = \mathbb E_i(\mathcal L(\vx_i)-\hat\mu(\mathcal L))^2 .
	\end{split}
\end{equation*}
A sample $\vx$ is considered an outlier if, and only if, $\mathcal L(\vx)$\,$<$\,$\hat\mu(\mathcal L)$\,$-$\,$c\sqrt{\hat\Sigma^2(\mathcal L)}$, where $c$ is a free parameter. 
Smaller values of $c$ will detect more outliers and vice versa. 
\subsection{Unconditional Sampling}\label{sec:gmr-sampling}
\noindent Sampling is, again, conducted according to GMM standard procedures. 
It consists of drawing a GMM component from a multinomial distribution parameterized by the GMM weights $\vpi$: $k$\,$\sim$\,$\mathcal{M}(\vpi)$. 
Then, a random vector $\vz$\,$\in$\,$\mathbb R^d$, $\vz$\,$\sim$\,$\mathcal{N}(0,\mI)$ of the same dimensions $d$ as the data is drawn.
The vector is transformed into a sample $\vx$ as $\vx$\,$=$\,$\mSigma_k \vz$\,$+$\,$\vmu_k$, which ensures that $\vx$\,$\sim$\,$\mathcal{N}_k(\cdot; \vmu_k, \mSigma_k)$.
In \cref{sec:disc:proof}, we will prove that the GMM log-likelihood on training data provides a lower bound for the log-likelihood of samples generated in this way. 
Thus, if we have higher training log-likelihoods, we can expect to generate better samples.
To show this, we prove the following.

\parheading{Proposition:}
The training loss of a GMM is a lower bound on the expected loss of generated samples. 
\parheading{Proof:} 
To prove the proposition, it is sufficient to prove the proposition for the case of a single Gaussian component density, which shall be denoted $\mathcal{N}(\vx; \vmu, \mSigma)$\,$\equiv$\,$\mathcal N(\vx)$. 
After decomposing the covariance matrix $\mSigma$ as $\mSigma$\,$=$\,$\mA\mA^\top$, a set of samples $G$\,$\supset$\,$\vg$ can be obtained (see \cref{sec:gmr-sampling}).
This is achieved by transforming a random normal variable $\vz$\,$\sim$\,$\mathcal N(0,\mI)$ as $\vg$\,$=$\,$A\vz$\,$+$\,$\vmu$. 
The loss on the generated samples is expressed as
\begin{align}
	\mathcal{L}(\vg) & = \ln \mathcal N(\vg) = \ln \mathcal N (A\vz + \vmu)                                         \nonumber      \\
                     & \sim f(\mSigma) - \frac{1}{2}(A\vz)^\top\mSigma^{-1}(A\vz) = f(\mSigma)-\frac{1}{2}\|\vz\|^2 \nonumber      \\
	\mathcal{L}(G)   & = \sum_i \mathcal{L}(\vg) = Nf(\mSigma) - \frac{dN}{2}.                                      \label{eqn:gen}
\end{align}
If the training samples follow a Gaussian distribution, their mean and variance coincide with the parameters $\vmu$, $\mSigma$ of the Gaussian component density.
Thus, their loss is identical to \cref{eqn:gen} by the same reasoning.
If training samples deviate from Gaussianity, as may be expected in practice, their loss will be lower. 
This is trivial to show by expanding their distribution around a Gaussian one into an Edgeworth series (see~\cite{blinnikov1998expansions}), and plugging this expansion into \cref{eqn:gen}. 
Hence, we know that the loss that is actually obtained on test data representing a lower bound for the loss of generated samples. 
\subsection{Class-Conditional Sampling} \label{sec:gmr-condsampling}
\noindent This form of sampling has the goal of generating samples belonging to a given class $c$.
To provide the GMM with this information, we fix a certain output vector $\vo$ of the linear classifier and try to infer which inputs $\vi$ would produce it:
\begin{equation*}
	\vo = \vs(\mW \vi  + \vb) \Rightarrow \vi \approx \mW^\top\big(\vs^{-1}(\vo) + \vk - \vb\big).
\end{equation*}
Since the softmax function is shift-invariant, the inverse is defined only up to a constant $\vk$ which we set to $0$. 
As a first approximation, we assume that the weight matrix $\mW$ of the linear classifier has orthogonal columns. 
Entries of $\vo$ must be bounded in the $[0,1]$ interval, have a unit sum, and express a confident decision for a given class $C$. 
We choose $o_C$\,$=$\,$0.95$ and normalize accordingly to obtain a \textit{control signal} $\vi$ for the GMM. 
This control signal represents the expected posterior probabilities of the GMM for a given class $C$. 
It is, therefore, consistent to use it for unconditional sampling (previous paragraph) instead of the GMM \mbox{weights $\vpi$}.

\subsection{Replay} \label{sec:gmr-replay}
\noindent Prior to training generator and learner at sub-task $T$\,$\ge$\,$1$, samples from previous sub-tasks $1$\,$\le$\,$t$\,$<$\,$T$ must be produced by the generator. 
If we let $\nu(t)$ denote the number of data samples for any sub-task $t$, and $\xi(t)$ the number of samples to generate for sub-task $t$, then two strategies may be discerned for choosing $\xi(t)$. 
The \textit{proportional} strategy which chooses $\xi(t)$\,$=$\,$\sum_{t'}^{t-1} \nu(t')$ and the \textit{constant} strategy with $\xi(t)$\,$=$\,$\kappa \nu(t)$.
\parheading{Training} 
Once samples have been generated, generator and learner are trained concurrently, each with its own loss function. 
For the GMM, we use plain SGD to maximize the \textit{log-likelihood} of the training data under the model, expressed in the notation of \cref{sec:gmr_intro} as:
\begin{equation*}
		\mathcal{L}(X) = \ln p(\mX)= \sum_i \log \sum_k \pi_j \mathcal N_j(\vx),
\end{equation*}
using the efficient training procedure for high-dimensional streaming data described in \cite{Gepperth2019}. 
The linear classifier receives the GMM responsibilities $\vgamma$ as input and is trained by minimizing the usual cross-entropy loss 
\begin{equation*}
	\begin{split}
		\vy_i           & = \vs\big(\mW\vgamma_i(\vx) + \vb\big) \\
		\mathcal L^{CE} & = \frac{1}{N}\sum_i \log y_{ij} t_{ij}
	\end{split}
\end{equation*}
by SGD, with $\vs(\cdot)$ denoting the softmax function. 
Learning rates are defined by $\epsilon^G$ for GMM and $\epsilon^C$ for the classifier. 
\parheading{Hyper-Parameters} 
The principal hyper-parameters of GMR are, first of all, the number $K$ of GMM components, and the GMM learning rate $\epsilon^{\text{GMM}}$. 
All GMM hyper-parameters are selected according to \cite{Gepperth2019}. 
In particular, the crucial parameter $K$ follows a \enquote{the more the better} logic so it is easy to select. 
For the linear classifier, the learning rate $\epsilon^{\text{C}}$ plays a role, as well. 
Since inputs to the linear classifier are normalized and bounded in the $[0,1]$ interval, the optimal learning rate is rather task-independent and it can be selected as a function of the GMM parameter $K$. 
\section{Elastic Weight Consolidation}\label{sec:EWC}
\noindent The approach from \cite{Kirkpatrick2016} is a typical regularization-based model for DNNs, see \cref{sec:relwork}.
EWC stores DNN parameters $\vec{\theta^{T_t}}$ after training on sub-task $T_t$.
In addition, EWC computes the \enquote{importance} of each parameter after training on sub-task $T_t$. 
This is done by approximating the diagonal  $\vec F^{T_t}$ of the Fisher Information Matrix (see \cite{Gepperth2019SIM} for a discussion of this approximation). 
The EWC loss function contains additional terms, see \cref{eqn:ewc_loss} besides the cross-entropy loss computed on the current sub-task $T_c$. 
These additional terms punish deviations from \enquote{important} DNN parameter values obtained after training on past sub-tasks:
\begin{equation}
	\mathcal{L}^{EWC} = \mathcal{L}_{T_c}(\theta) + \frac{\lambda}{2}\sum_{t=1}^{c-1} \sum_i F^{T_t}_i\Big(\theta_i - \theta^{T_t}_i\Big)^2 
	\label{eqn:ewc_loss}
\end{equation} 
EWC is optimized using the Adam optimizer. 
EWC hyper-parameters are the SGD step size $\epsilon^{EWC}$, the regularization constant $\lambda$ and the number and size of layers in the DNN. 
In \cite{Kirkpatrick2016}, it is proposed to set $\lambda$\,$=$\,$1/\epsilon^{EWC}$, thereby eliminating one hyper-parameter.
\section{Generative Replay}\label{sec:gr}
\noindent We implement generative replay (GR) as described in \cite{Shin2017} with a GAN-based generator.
The precise configurations of the generator and learner is given in \cref{app:1}.
Batch sizes are $\mathcal{B}$ for the first sub-task and $2\mathcal{B}$ for sub-tasks $t$\,$>$\,$1$. 
Important hyper-parameters are the SGD step size $\epsilon^G$, and the number of epochs $\mathcal{E}$ for training. 
At each sub-task, the generator produces as many samples as contained in all previous sub-tasks to maintain balance. 
Alternatively, a fixed number of generated samples is possible, as well.
\section{Experiments}\label{sec:exp}
\noindent For validating the goals outlined in \cref{sec:intro}, we conduct the following experiments on sequential learning tasks (SLTs), which are constructed as described in \cref{sec:SLT}.
In \cref{sec:exp1}, we demonstrate unsupervised outlier detection to identify sub-task boundaries without reference to class labels. 
A demonstration of sampling quality as measured by the GMM-log-likelihood is given in \cref{sec:exp2}. 
As a by-product of the GMR architecture, we present results on class-conditional sampling on all three datasets in \cref{sec:exp3}.
\Cref{sec:exp4} shows that GMR achieves state-of-the-art classification performance on the SLTs when compared to generative replay and EWC. 
\subsection{Hyper-Parameters}\label{sec:hp}
\parheading{GMR} 
In terms of \cref{sec:gmr-details}, we chose $K$\,$=$\,$100$, $\epsilon^{G}$\,$=$\,$0.01$, $\mathcal{B}$\,$=$\,$100$, $\epsilon^{C}$\,$=$\,$0.01$.
For the constant replay strategy (see \cref{sec:gmr-replay}), a proportionality constant of $\kappa$\,$=$\,$2$ is used.
Training epochs are empirically set to $50$ for $T_1$.
For all other sub-tasks, the number was doubled compared to the previous one.
The other hyper-parameters are set to default values as defined in \cite{Gepperth2019}. 
\parheading{GR}
In terms of \cref{sec:gr} and \cref{fig:replay}, generators are always trained for $50$ epochs ($\mathcal{E}$) and solvers for $25$ epochs. 
The Adam optimizer is used for effecting gradient descent, using a step size $\epsilon^{GR}$\,$=$\,$0.001$ for solver and generators.
Samples are generated in order to maintain balance between previous and current classes.
\parheading{EWC}
For each SLT, we perform a grid search for the parameter $\epsilon$.
We vary the learning rate for EWC $\epsilon^{EWC}$ as $\epsilon^{EWC}$\,$\in$\,$\{0.001,0.0001,0.00001,0.000001,0.0000001\}$, depending on $\lambda$ being always set to $\frac{1}{\epsilon^{EWC}}$.
We fix the model architecture to a three-layer DNN, each of the size $800$.
Training epochs $\mathcal{E}$ are empirically set to $10$ for each training task. 
The best hyper-parameters and experiments are selected based on the highest average accuracy on all classes measured during the training process.
\subsection{Task Boundary Detection}\label{sec:exp1} 
\begin{figure}[htb!]
	\centering
	\includegraphics[width=0.8\linewidth]{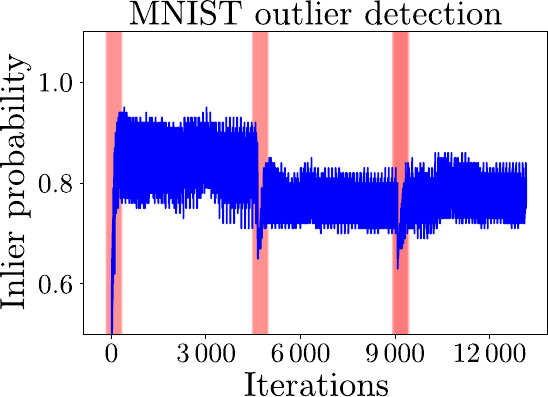} \\
	\vspace{1em}
	\includegraphics[width=0.8\linewidth]{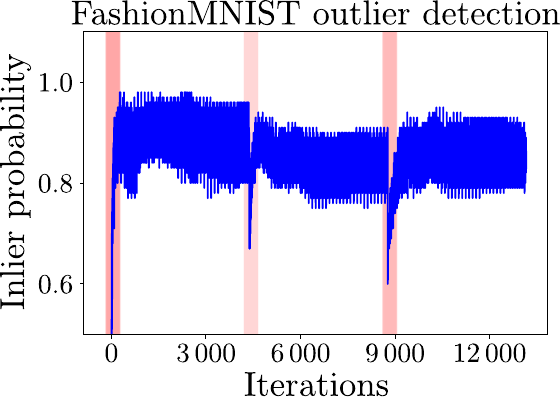} 
	\caption{
		Detection of task boundaries on the first three sub-tasks of $D_{2\text{-}2\text{-}2\text{-}2\text{-}2a}$.
		Areas highlighted in red signal automatically detected sub-task boundaries.
	}
	\label{fig:taskboundaries}
\end{figure}

\noindent We train GMR on the SLT $D_{2\text{-}2\text{-}2\text{-}2\text{-}2a}$, while updating the sliding average and variance for the log-likelihood as indicated in \cref{sec:gmr-outliers}. 
For each sample $\vx$ in a mini-batch $\mathcal{B}$, we test whether they are inliers as discussed in \cref{sec:gmr-outliers}, using a value of $c$\,$=$\,$1$. 
We then compute the empirical probability of inliers in the mini-batch. 
Each time the probability drops by more than $20\%$, we assume a sub-task boundary has occurred. 
The results are shown in \cref{fig:taskboundaries}.
\subsection{Sampling}\label{sec:exp2} 
\begin{figure}[htb!]
	\centering
	\includegraphics[width=0.8\linewidth]{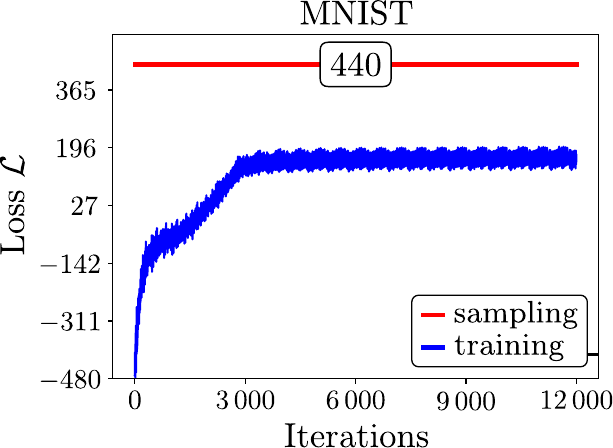} \\
	\vspace{1em}
	\includegraphics[width=0.8\linewidth]{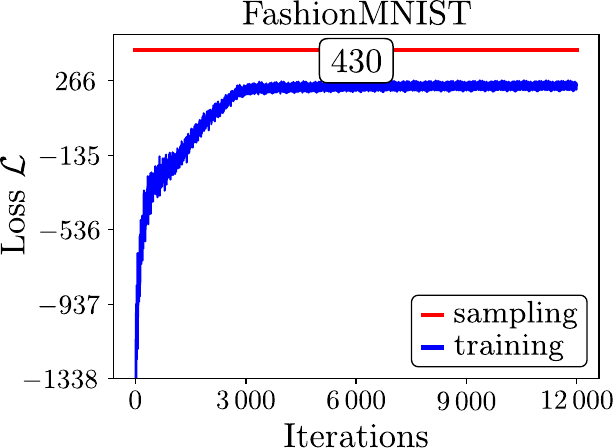}
	\caption{
		Training and sampling loss for SLT $D_{10}$.
		The sampling loss is superimposed on each graph as a red horizontal line. 
		Its value is given in the box. 
	}
	\label{fig:sampling}
\end{figure}

\noindent In this experiment, we verify that the GMM loss of generated samples (sampling loss) is always higher than the GMM training loss. 
A proof for this was given in \cref{sec:gmr-sampling}.
Here, we give an empirical validation.
This experiment is independent of continual learning, which is why we use the baseline SLT $D_{10}$. 
\Cref{fig:sampling} shows the results for two datasets, and we observe that the sampling loss is indeed higher than the asymptotic training loss, often by a quite large margin.
\subsection{Class-Conditional Sampling with GMR}\label{sec:exp3}
\noindent For this experiment, we train a GMR instance on SLT $D_{10}$ for each dataset, i.e., on all classes at once. 
Subsequently, we use each of the three trained models to conditionally generate $50$ samples: $25$ from classes $\mathcal{C}$\,$=$\,$\{1,2\}$, and $25$ samples from classes $\mathcal{C}$\,$=$\,$\{5,7\}$. 
For each generated sample, the class $c$ is drawn from $\mathcal{C}$ with equal probability.
Control signals to the GMM for generating a sample from class $c$ are obtained and applied according to \cref{sec:gmr-condsampling}. 
The results can be viewed in \cref{fig:condsampling}. 
We observe that samples are reliably selected from the given set $\mathcal{C}$.
In some cases, errors occur for samples that are visually very similar to elements of $\mathcal{C}$.
This reflects the fact that the classification accuracy is not perfect. 
For perfect classification, we expect no such sampling inaccuracies.
\newcommand\myimagefactor{0.36}
\newcommand\myimagefactorS{0.39}
\begin{figure}[htb!]
	\centering
	\includegraphics[width=\myimagefactor\linewidth]{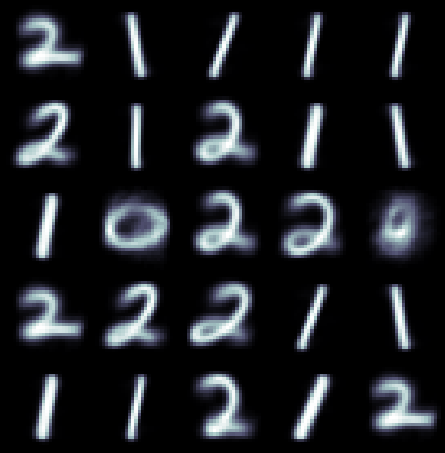}\hspace{1em}\includegraphics[width=\myimagefactor\linewidth]{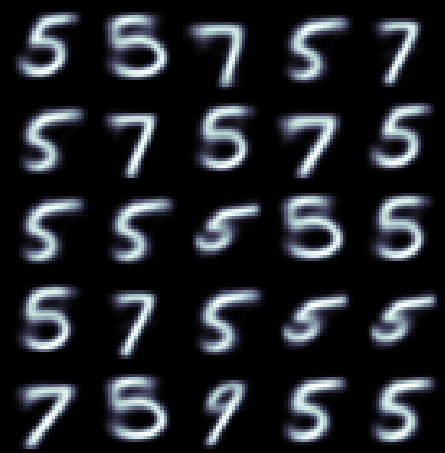}\\\vspace{.5em}
	\includegraphics[width=\myimagefactor\linewidth]{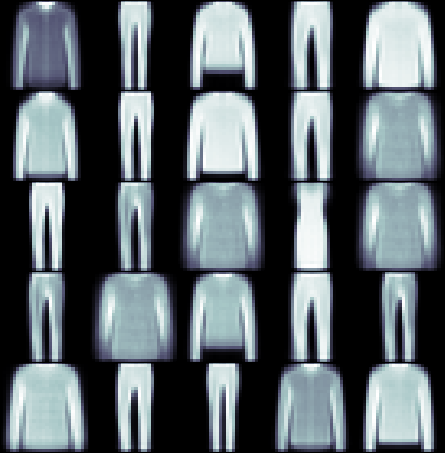}\hspace{1em}\includegraphics[width=\myimagefactor\linewidth]{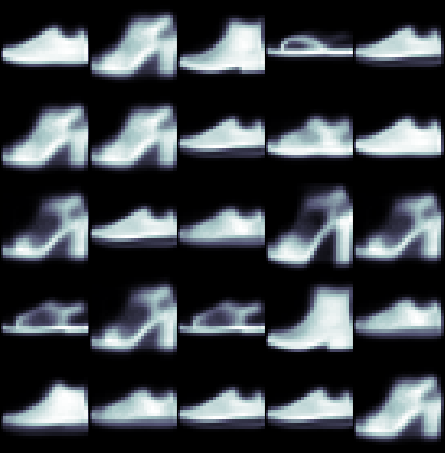}\\\vspace{.5em}
	\includegraphics[width=\myimagefactor\linewidth]{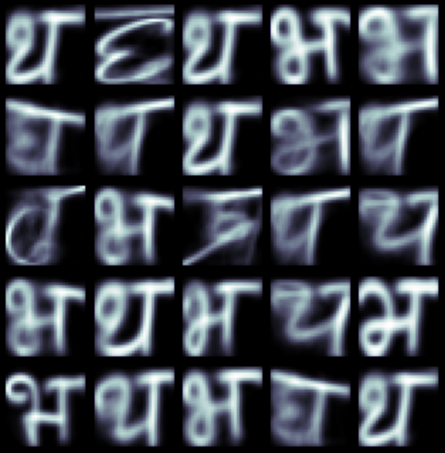}\hspace{1em}\includegraphics[width=\myimagefactor\linewidth]{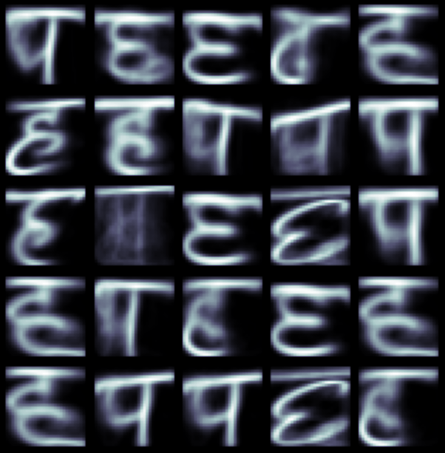} 
	\caption{
		Conditional sampling results for GMR models trained on MNIST (top), FashionMNIST (middle) and Devanagari (bottom). 
		In each row, $25$ samples for classes 1,2 (left) and $25$ samples for classes 5,7 (right) are generated. 
	}
	\label{fig:condsampling}
\end{figure}
\par
\begin{figure}[htb!]
	\centering
	\includegraphics[width=\myimagefactorS\linewidth]{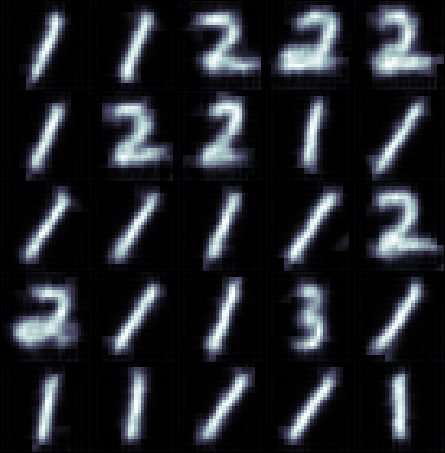}\hspace{1em}\includegraphics[width=\myimagefactorS\linewidth]{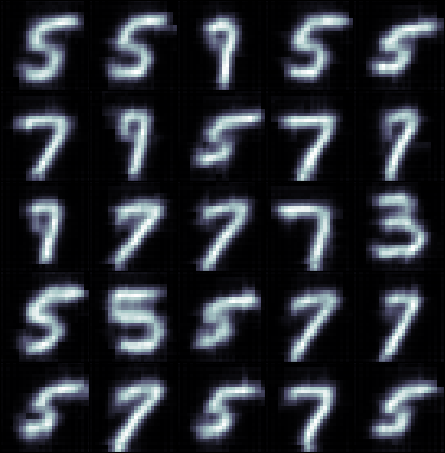}
	\caption{
		Conditional sampling results for GMR models trained on MNIST using a deep convolutional GMM.  
		To be read as \cref{fig:condsampling}.
	}
	\label{fig:dcgmm}
\end{figure}

Additionally, we perform class-conditional sampling in the same way as described in \cite{gepperth2021a}, but use a deep convolutional GMM (DCGMM).
Model details are given in \cref{app:2}. 
\Cref{fig:dcgmm} shows the generated samples for MNIST.
\subsection{Comparison of GR, GMR and EWC}\label{sec:exp4}
\begin{table*}[htb]
	\footnotesize
	\newlength{\mycolumnwidth}
	\setlength{\mycolumnwidth}{20.5pt}
	\newcolumntype{b}{>{\centering\arraybackslash}m{\mycolumnwidth}}
	\centering
	\caption{Results of the conducted GMR, EWC and GR experiments. 
		The accuracy in $\%$ is stated as baseline for all experiments based on the available classes for each dataset.
		For each best SLT experiment (defined in \cref{tab:SLTs}) the difference to the baseline is given.
		Therefore, the maximum measured accuracy value is used, averaged over $10$ experiment repetitions.
		To measure the accuracy, the joint test dataset consisting of all tasks ($D_{10}$) is used. 
	}
	\label{tab:results}
	\setlength\tabcolsep{2pt} 
	\begin{tabular}{|l|bb|bb|bb!{\vrule width 1pt}bb|bb|bb!{\vrule width 1pt}bb|bb|bb|}
		\hline
		\multirow{3}{*}{\diagbox[width=44pt,height=28.5pt]{SLT}{\shortstack{\textbf{model}\\{\scriptsize dataset}}}} &                                                                                   \multicolumn{6}{c!{\vrule width 1pt}}{\textbf{GMR}}                                                                                   &                                                                                    \multicolumn{6}{c!{\vrule width 1pt}}{\textbf{EWC}}                                                                                    &                                                                                            \multicolumn{6}{c|}{\textbf{GR}}                                                                                            \\
		                                                                                                             &                       \multicolumn{2}{c|}{MNIST}                       &            \multicolumn{2}{c|}{{\scriptsize FashionMNIST}}            &           \multicolumn{2}{c!{\vrule width 1pt}}{Devanagari}            &                       \multicolumn{2}{c|}{MNIST}                       &            \multicolumn{2}{c|}{{\scriptsize FashionMNIST}}             &            \multicolumn{2}{c!{\vrule width 1pt}}{Devanagari}            &                      \multicolumn{2}{c|}{MNIST}                       &            \multicolumn{2}{c|}{{\scriptsize FashionMNIST}}            &                    \multicolumn{2}{c|}{Devanagari}                     \\
		                                                                                                             & acc.\,{\scriptsize $\%$}           & std                               & acc.\,{\scriptsize $\%$}          & std                               & acc.\,{\scriptsize $\%$}           & std                               & acc.\,{\scriptsize $\%$}           & std                               & acc.\,{\scriptsize $\%$}           & std                               & acc.\,{\scriptsize $\%$}           & std                                & acc.\,{\scriptsize $\%$}           & std                              & acc.\,{\scriptsize $\%$}           & std                              & acc.\,{\scriptsize $\%$}           & std                               \\ \hline
		$D_{10}$ {\scriptsize baseline}                                                                              & \tablenum[table-format=2.1]{87.4}  & \tablenum[table-format=1.2]{0.59} & \tablenum[table-format=2.1]{73.9} & \tablenum[table-format=1.2]{0.26} & \tablenum[table-format=2.1]{74.1}  & \tablenum[table-format=1.2]{0.73} & \tablenum[table-format=2.1]{97.57} & \tablenum[table-format=1.2]{0.26} & \tablenum[table-format=2.1]{87.55} & \tablenum[table-format=1.2]{0.38} & \tablenum[table-format=2.1]{95.58} & \tablenum[table-format=2.2]{0.56}  & \tablenum[table-format=2.1]{99.3}  & \tablenum[table-format=1.2]{0}   & \tablenum[table-format=2.1]{99.3}  & \tablenum[table-format=1.2]{0}   & \tablenum[table-format=2.1]{99.1}  & \tablenum[table-format=2.2]{0}    \\ \hline
		                                                                                                             & diff.                              & std                               & diff.                             & std                               & diff.                              & std                               & diff.                              & std                               & diff.                              & std                               & diff.                              & std                                & diff.                              & std                              & diff.                              & std                              & diff.                              & std                               \\ \hline
		$D_{9\text{-}1a}$                                                                                            & \tablenum[table-format=3.1]{-1.3}  & \tablenum[table-format=1.2]{0.59} & \tablenum[table-format=3.1]{-2.7} & \tablenum[table-format=1.2]{0.26} & \tablenum[table-format=3.1]{-3.2}  & \tablenum[table-format=1.2]{0.73} & \tablenum[table-format=3.1]{-41.8} & \tablenum[table-format=1.2]{0.26} & \tablenum[table-format=3.1] {-9.6} & \tablenum[table-format=1.2]{0.38} & \tablenum[table-format=3.1]{-56.6} & \tablenum[table-format=2.2]{0.56}  & \tablenum[table-format=3.1]{-15.1} & \tablenum[table-format=1.2]{0.7} & \tablenum[table-format=3.1]{-25.6} & \tablenum[table-format=1.2]{0.5} & \tablenum[table-format=3.1]{-13.5} & \tablenum[table-format=2.2]{0.54} \\ \hline
		$D_{9\text{-}1b}$                                                                                            & \tablenum[table-format=3.1]{-3.5}  & \tablenum[table-format=1.2]{2.19} & \tablenum[table-format=3.1]{-1.5} & \tablenum[table-format=1.2]{0.87} & \tablenum[table-format=3.1]{-1.4}  & \tablenum[table-format=1.2]{0.88} & \tablenum[table-format=3.1]{-50.7} & \tablenum[table-format=1.2]{7.77} & \tablenum[table-format=3.1]{-20.1} & \tablenum[table-format=1.2]{2.52} & \tablenum[table-format=3.1]{-29.7} & \tablenum[table-format=2.2]{13.34} & \tablenum[table-format=3.1]{-21.8} & \tablenum[table-format=1.2]{0.9} & \tablenum[table-format=3.1]{-16.5} & \tablenum[table-format=1.2]{1.1} & \tablenum[table-format=3.1]{-10.9} & \tablenum[table-format=2.2]{0.41} \\ \hline
		$D_{5\text{-}5a}$                                                                                            & \tablenum[table-format=3.1]{-0.6}  & \tablenum[table-format=1.2]{1.53} & \tablenum[table-format=3.1]{-1.2} & \tablenum[table-format=1.2]{1.53} & \tablenum[table-format=3.1]{-6.8}  & \tablenum[table-format=1.2]{1.38} & \tablenum[table-format=3.1]{-35.3} & \tablenum[table-format=1.2]{6.65} & \tablenum[table-format=3.1]{-32.7} & \tablenum[table-format=1.2]{4.22} & \tablenum[table-format=3.1]{-46.0} & \tablenum[table-format=2.2]{15.38} & \tablenum[table-format=3.1]{-10.0} & \tablenum[table-format=1.2]{1.4} & \tablenum[table-format=3.1]{-19.8} & \tablenum[table-format=1.2]{3.2} & \tablenum[table-format=3.1]{-6.7}  & \tablenum[table-format=2.2]{0.3}  \\ \hline
		$D_{5\text{-}5b}$                                                                                            & \tablenum[table-format=3.1]{-1.3}  & \tablenum[table-format=1.2]{1.92} & \tablenum[table-format=3.1]{-1.9} & \tablenum[table-format=1.2]{0.49} & \tablenum[table-format=3.1]{-4.7}  & \tablenum[table-format=1.2]{1.59} & \tablenum[table-format=3.1]{-35.0} & \tablenum[table-format=1.2]{1.83} & \tablenum[table-format=3.1]{-36.0} & \tablenum[table-format=1.2]{2.72} & \tablenum[table-format=3.1]{-47.1} & \tablenum[table-format=2.2]{0.11}  & \tablenum[table-format=3.1]{-11.9} & \tablenum[table-format=1.2]{1.0} & \tablenum[table-format=3.1]{-17.7} & \tablenum[table-format=1.2]{4.0} & \tablenum[table-format=3.1]{-7.6}  & \tablenum[table-format=2.2]{0.37} \\ \hline
		$D_{2\text{-}2\text{-}2\text{-}2\text{-}2a}$                                                                 & \tablenum[table-format=3.1]{-9.5}  & \tablenum[table-format=1.2]{3.83} & \tablenum[table-format=3.1]{-8.5} & \tablenum[table-format=1.2]{0.91} & \tablenum[table-format=3.1]{-22.5} & \tablenum[table-format=1.2]{2.71} & \tablenum[table-format=3.1]{-72.2} & \tablenum[table-format=1.2]{7.43} & \tablenum[table-format=3.1]{-55.6} & \tablenum[table-format=1.2]{4.05} & \tablenum[table-format=3.1]{-72.1} & \tablenum[table-format=2.2]{2.75}  & \tablenum[table-format=3.1]{-41.4} & \tablenum[table-format=1.2]{3.8} & \tablenum[table-format=3.1]{-25.0} & \tablenum[table-format=1.2]{5.9} & \tablenum[table-format=3.1]{-40.0} & \tablenum[table-format=2.2]{3.3}  \\ \hline
		$D_{2\text{-}2\text{-}2\text{-}2\text{-}2b}$                                                                 & \tablenum[table-format=3.1]{-10.4} & \tablenum[table-format=1.2]{5.28} & \tablenum[table-format=3.1]{-5.7} & \tablenum[table-format=1.2]{2.37} & \tablenum[table-format=3.1]{-14.7} & \tablenum[table-format=1.2]{2.94} & \tablenum[table-format=3.1]{-72.6} & \tablenum[table-format=1.2]{3.22} & \tablenum[table-format=3.1]{-57.3} & \tablenum[table-format=1.2]{4.99} & \tablenum[table-format=3.1]{-73.2} & \tablenum[table-format=2.2]{2.31}  & \tablenum[table-format=3.1]{-34.8} & \tablenum[table-format=1.2]{4.1} & \tablenum[table-format=3.1]{-29.4} & \tablenum[table-format=1.2]{7.3} & \tablenum[table-format=3.1]{-34.7} & \tablenum[table-format=2.2]{7.6}  \\ \hline
	\end{tabular} 
\end{table*}
\noindent We train EWC, GMR and GR on all SLTs listed in \cref{sec:SLT}, according to the hyper-parameter settings described earlier, see \cref{sec:hp}. 
Classification accuracy is read off after completing training on the last sub-task.
Baseline accuracy on a non-continual learning task ($D_{10}$) is recorded for all datasets and methods. 
For GR and GMR, we use the proportional sample generation strategy, see \cref{sec:gmr-replay}.
We present the results as deviations from baseline performance in \cref{tab:results}. 
The comparison is not entirely fair since GMR has a much lower baseline performance. 
On the contrary, we observe that the drop in classification accuracy due to CL is generally much smaller.
We interpret this minor drop as characterization of continual learning performance.
\section{Principal Conclusions and Discussion}
\subsection{State-Of-The-Art GMR Performance}
\noindent From the experiments of \cref{sec:exp}, we can conclude that GMR can egalize the performance of GR, and that both GMR and GR outperform EWC by a large margin. 
Here, we are talking about \textit{continual learning} performance as defined in \cref{sec:exp4}. 
GMR performance on the non-continual baseline $D_{10}$ is inferior to that of a standard DNN. 
This makes it even more remarkable that continual learning performance is similar to GR, which included a fully-fledged CNN.
\subsection{Memory Requirements}
\noindent GMR has a low memory footprint because the generator (the GMM) is re-used for classification, see \cref{fig:gmr}. 
Since the GMM itself is \enquote{flat}, its memory requirements are modest. 
For an input dimensionality of $d$\,$=$\,$1000$, with $K$\,$=$\,$100$ GMM components and $10$ classes, the total memory required to store the complete GMR model is $2Kd$\,$+$\,$10K$\,$+$\,$10$\,$=$\,$201$\,$010$. 
The corresponding GR model consists of a three-layer DNN (the learner) and two CNNs for the generator and a discriminator. 
It contains $3$\,$770$\,$204$ parameters, which is more than two orders of magnitude larger than the corresponding GMR model.
This is remarkable as continual learning performance is comparable. 
\subsection{Quality of Generated Samples}\label{sec:disc:proof}
\noindent In contrast to models like, e.g., GANs, GMMs provide strong guarantees concerning the quality of generated samples via their loss function, see \cref{sec:gmr-sampling}.
An implication is that sample generation capacity can be monitored \textit{at training time} by observing the loss. 
If the loss decreases significantly during training, this would be a strong sign for mode collapse. 
This is virtually excluded due to SGD that aims to maximize the loss, and such behavior was never observed in all experiments conducted in \cref{sec:exp2}.
\subsection{Simple Conditional Sampling} 
\noindent As shown in \cref{sec:exp3}, conditional sampling is a reliable way to obtain samples from certain given classes. 
The simplicity of the process is an appealing feature of GMR, realized through the GMMs ability for unconditional sampling.
\subsection{Respecting Real-World Constraints}
\noindent GMR is attractive for real-world applications because it fully respects several important constraints (see \cite{Pfuelb2019} for a more comprehensive discussion of real-world constraints). 
\parheading{No Looking Back} 
GMR does not require access to data from past sub-tasks to determine when to stop training, see \cref{sec:intro}.
This property is shared with GR but not with EWC, whose performance drops after a certain time, see \cref{sec:exp4}.
To determine the optimal point for stopping the training, EWC requires access to data from past sub-tasks.
However, it is a contradiction to the continual learning paradigm.
\parheading{No Looking Ahead} 
The hyper-parameters for EWC, notably the balancing parameter $\lambda$ and the DNN parameters, need to be determined by grid-search, since performance depends upon these parameters in a complex way.
This requires repeating the whole experiment many times with different parameters, and thus determining hyper-parameters for given sub-task based on future sub-tasks.
That violates the CL paradigm, which states that only one sub-task at a time can be accessed. 
See \cite{Pfuelb2019} for a longer discussion on this point. 
In contrast, GMR has only one free parameter: the number of GMM components $K$, which follows a \textit{the-more-the-better}. 
Therefore, it is possible to determine a good value for $K$ on sub-task $T_1$ only, and thus to respect the CL paradigm. 
A large number of training epochs does not affect learning adversely, and can be selected on $T_1$ more liberally, similar to the learning rate. 
\subsection{Model Limitations}
\noindent As far as \textit{continual} learning performance is concerned, the presented GMR method leads to state-of-the-art performance on SLTs derived from simple benchmarks such as MNIST or FashionMNIST. 
It is, however, strongly inferior w.r.t.\ \textit{non-continual} (baseline) performance as shown in \cref{sec:exp4}.
Neither can it be expected to perform well on more difficult SLTs constructed, e.g., from the SVHN benchmark. 
Most importantly, such a complex dataset would require an extremely high number of GMM components for high-quality sampling. 
In addition, the representation provided by a flat GMM may not be expressive enough to allow accurate classification.
A solution to both problems could be the use of deep GMM variants as presented in \cite{VanDenOord2014} or \cite{gepperth2021a}.
\section{Outlook and Next Steps}
\noindent Applying GMR to more challenging problems requires the improvement of the GMM's sample generation capacity without excessive resource requirements. 
We will investigate two main directions:
\parheading{Using a Deep Convolutional Generator}
Just as DNNs and CNNs can represent more complex functions than single-layer perceptrons, deep convolutional GMMs can model more complex distributions. 
We plan to investigate the models proposed in \cite{VanDenOord2014} or \cite{gepperth2021a} for replacing the current flat GMM by a deep convolutional variant.
\parheading{Different GMR Design Choices}
A major design choice in GMR is to restrict GMMs to diagonal covariance matrices, see \cref{sec:gmr-details}. 
Full covariance matrices are out of the question due to memory reasons: for $K$\,$=$\,$100$ and data dimensionality $d$\,$\approx$\,$1000$.
This would involve $10^8$ parameters for storage alone, not to mention memory requirements on a GPU due to parallel processing.
A compromise might be the use of a Mixture of Factor Analyzers (MFA) instead of a GMM model.
With a reasonable memory overhead this may significantly enhance the GMM's sample generation capacity as demonstrated in, e.g., \cite{Richardson2018}.
\bibliographystyle{unsrt}
\bibliography{egbib,cl2021}
\clearpage
\appendix
\section{Generative Replay: Model Details}\label{app:1}
\noindent The solver component in generative replay is implemented by a DNN, whose parameters are given in \cref{tab:app1}.
For the generator, we chose a Generative Adversarial Network (GAN). 
Generator/discriminator configuration for GANs is given in \cref{tab:app3}.
\begin{table}[htb!]
\centering
\caption{
	Solver configuration and number of parameters for each layer. 
	$\mathcal{B}$ indicates the used mini-batch size. 
}
\label{tab:app1}
\begin{tabular}{|ccc|}
	\hline
	\textbf{Layer Type} &   \textbf{Output Shape}    & \textbf{Parameters}                 \\ \hline
	      Reshape       & ($\mathcal{B}$, 28, 28, 1) & \tablenum[table-format=6.0]{0}      \\
	      Flatten       &    ($\mathcal{B}$, 784)    & \tablenum[table-format=6.0]{0}      \\
	       Dense        &    ($\mathcal{B}$, 400)    & \tablenum[table-format=6.0]{314000} \\
	       ReLU         &    ($\mathcal{B}$, 400)    & \tablenum[table-format=6.0]{0}      \\
	       Dense        &    ($\mathcal{B}$, 400)    & \tablenum[table-format=6.0]{160400} \\
	       ReLU         &    ($\mathcal{B}$, 400)    & \tablenum[table-format=6.0]{0}      \\
	       Dense        &    ($\mathcal{B}$, 400)    & \tablenum[table-format=6.0]{160400} \\
	       ReLU         &    ($\mathcal{B}$, 400)    & \tablenum[table-format=6.0]{0}      \\
	       Dense        &    ($\mathcal{B}$, 10)     & \tablenum[table-format=6.0]{4010}   \\
	      Softmax       &    ($\mathcal{B}$, 10)     & \tablenum[table-format=6.0]{0}      \\ \hline
\end{tabular}
\end{table}
\begin{table}[htb!]
\centering
\caption{
	GAN generator/discriminator configuration and number of parameters for each layer. 
	$\mathcal{B}$ indicates the used mini-batch size.
}
\label{tab:app3}
\begin{tabular}{|ccc|}
	\hline
	\textbf{Layer Type} &    \textbf{Output Shape}    &                  \textbf{Parameters} \\ \hline
	\multicolumn{3}{|c|}{Generator}                                                          \\ \hline
	      Flatten       &     ($\mathcal{B}$, 80)     &       \tablenum[table-format=7.0]{0} \\
	       Dense        &   ($\mathcal{B}$, 12544)    & \tablenum[table-format=7.0]{1016064} \\
	     BatchNorm      &   ($\mathcal{B}$, 12544)    &   \tablenum[table-format=7.0]{50176} \\
	     LeakyReLU      &   ($\mathcal{B}$, 12544)    &       \tablenum[table-format=7.0]{0} \\
	      Reshape       & ($\mathcal{B}$, 7, 7, 256)  &       \tablenum[table-format=7.0]{0} \\
	    Conv2DTrans     & ($\mathcal{B}$, 7, 7, 128)  &  \tablenum[table-format=7.0]{819328} \\
	     BatchNorm      & ($\mathcal{B}$, 7, 7, 128)  &     \tablenum[table-format=7.0]{512} \\
	     LeakyReLU      & ($\mathcal{B}$, 7, 7, 128)  &       \tablenum[table-format=7.0]{0} \\
	    Conv2DTrans     & ($\mathcal{B}$, 14, 14, 64) &  \tablenum[table-format=7.0]{204864} \\
	       Batch        & ($\mathcal{B}$, 14, 14, 64) &     \tablenum[table-format=7.0]{256} \\
	     LeakyReLU      & ($\mathcal{B}$, 14, 14, 64) &       \tablenum[table-format=7.0]{0} \\
	    Conv2DTrans     & ($\mathcal{B}$, 28, 28, 1)  &    \tablenum[table-format=7.0]{1601} \\ \hline
	\multicolumn{3}{|c|}{Discriminator}                                                      \\ \hline
	      Reshape       & ($\mathcal{B}$, 28, 28, 1)  &       \tablenum[table-format=6.0]{0} \\
	      Conv2D        & ($\mathcal{B}$, 14, 14, 64) &    \tablenum[table-format=6.0]{1664} \\
	     LeakyReLU      & ($\mathcal{B}$, 14, 14, 64) &       \tablenum[table-format=6.0]{0} \\
	      Dropout       & ($\mathcal{B}$, 14, 14, 64) &       \tablenum[table-format=6.0]{0} \\
	      Conv2D        & ($\mathcal{B}$, 7, 7, 128)  &  \tablenum[table-format=6.0]{204928} \\
	     LeakyReLU      & ($\mathcal{B}$, 7, 7, 128)  &       \tablenum[table-format=6.0]{0} \\
	      Dropout       & ($\mathcal{B}$, 7, 7, 128)  &       \tablenum[table-format=6.0]{0} \\
	      Conv2D        & ($\mathcal{B}$, 7, 7, 256)  &  \tablenum[table-format=6.0]{819456} \\
	     LeakyReLU      & ($\mathcal{B}$, 7, 7, 256)  &       \tablenum[table-format=6.0]{0} \\
	      Dropout       & ($\mathcal{B}$, 7, 7, 256)  &       \tablenum[table-format=6.0]{0} \\
	      Flatten       &   ($\mathcal{B}$, 12544)    &       \tablenum[table-format=6.0]{0} \\
	       Dense        &     ($\mathcal{B}$, 1)      &   \tablenum[table-format=6.0]{12545} \\ \hline
\end{tabular}
\end{table}
\section{Deep convolutional GMMs: Details} \label{app:2}
\noindent In \cref{tab:app4}, we give details about the different layers in the deep convolutional GMM (DCGMM).
For the sampling experiments in \cref{sec:exp2}, we follow the terms and definitions of \cite{gepperth2021a}. 
The total number of parameters is $633\,298$. 
\begin{table}[htb!]
\setlength\tabcolsep{4pt}
\centering
\caption{
	DCGMM configuration and number of parameters for each layer. 
	$\mathcal{B}$ indicates the used mini-batch size.
}
\label{tab:app4}
\begin{tabular}{|cccc|}
	\hline
	\multirow{2}{*}{\textbf{Layer Type}} &   \textbf{Filter-}   & \multirow{2}{*}{\textbf{Output Shape}} & \multirow{2}{*}{\textbf{Parameters}} \\
	                                     & \textbf{Size/Stride} &                                        &                                      \\ \hline
	              Folding                &        8 / 2         &      ($\mathcal{B}$, 11, 11, 64)       &    \tablenum[table-format=6.0]{0}    \\
	                GMM                  &          -           &      ($\mathcal{B}$, 11, 11, 81)       &  \tablenum[table-format=6.0]{5184}   \\
	              Folding                &        11 / 1        &      ($\mathcal{B}$, 1, 1, 7744)       &    \tablenum[table-format=6.0]{0}    \\
	                GMM                  &          -           &       ($\mathcal{B}$, 1, 1, 81)        & \tablenum[table-format=6.0]{627264}  \\
	             Classifier              &          -           &          ($\mathcal{B}$, 10)           &   \tablenum[table-format=6.0]{810}   \\ \hline
\end{tabular}
\end{table}

\newcommand\blfootnote[1]{%
	\begingroup
	\renewcommand\thefootnote{}\footnote{#1}%
	\addtocounter{footnote}{-1}%
	\endgroup
}
\blfootnote{Accepted as Findings at the CLVISION2021 workshop.} 
\end{document}